\documentclass{article}

\usepackage[nonatbib, main, final]{neurips_2026}
\usepackage[sort,comma,numbers]{natbib}

\usepackage[utf8]{inputenc} % allow utf-8 input
\usepackage[T1]{fontenc}    % use 8-bit T1 fonts
\usepackage{hyperref}       % hyperlinks
\hypersetup{
    colorlinks,
    linkcolor={red!50!black},
    citecolor={blue!50!black},
    urlcolor={blue!80!black}
}
\usepackage{amsmath}
\usepackage{cleveref}
\usepackage{url}            % simple URL typesetting
\usepackage{booktabs}       % professional-quality tables
\usepackage{amsfonts}       % blackboard math symbols
\usepackage{nicefrac}       % compact symbols for 1/2, etc.
\usepackage{microtype}      % microtypography
\usepackage{xcolor}         % colors
\usepackage{ulem}           % barrer du texte
\usepackage{float}

\DeclareMathOperator*{\argmax}{arg\,max}

\usepackage{enumitem}

\newcommand{\bmM}{\mathcal{M}}
\newcommand{\bmS}{\mathcal{S}}

\newcommand{\bmZ}{\mathcal{Z}}

\newcommand{\textoverline}[1]{$\overline{\mbox{#1}}$}

\newcommand{\msalignplus}{\textoverline{MSAlign}}

\definecolor{darkgreen}{rgb}{0.1,0.5,0.1}
\definecolor{darkblue}{rgb}{0.1,0.1,0.6}

\title{MSAlign: Aligning Molecule and Mass Spectra \\ Representations for Metabolite Identification}

\author{%
  Paul~Krzakala \\
  LTCI, Télécom Paris \& CMAP, Ecole Polytechnique,\\ 
  Institut Polytechnique de Paris
  \AND
  Gabriel~Melo \\
  LTCI, Télécom Paris, \\ 
  Institut Polytechnique de Paris
  \And 
  Camille~Lançon \\
  CEA, INRAE, MetaboHUB, \\ 
  Université Paris-Saclay 
  \And
  Charlotte~Laclau \\
  LTCI, Télécom Paris, \\ 
  Institut Polytechnique de Paris
  \AND
  Rémi~Flamary \\
  CMAP, Ecole Polytechnique, \\ 
  Institut Polytechnique de Paris
  \And
  Etienne~Thévenot \\
  CEA, INRAE, MetaboHUB, \\ 
  Université Paris-Saclay
  \And
  Florence~d'Alché{-}Buc \\
  LTCI, Télécom Paris, \\ 
  Institut Polytechnique de Paris
}
\usepackage{packages}
\begin{document}

\maketitle

\begin{abstract}
Accurately identifying metabolites i.e. small molecules from mass spectrometry
data remains a core challenge in metabolomics, with broad applications in drug
discovery, environmental analysis, and clinical research. We address the
Molecule Retrieval task, which consists in recovering the chemical structure of
a metabolite from its MS/MS spectrum given a set of candidate molecules. We make
three contributions. First, we propose a unified framework encompassing recent
approaches based on representation alignment and contrastive learning. Second,
we introduce MSAlign, a lightweight model that achieves state-of-the art
performances by aligning two frozen foundation models (DreaMS for mass spectra
and MolDeBERTa for molecules) and demonstrate that a score fusion strategy
further improves the performance for a very small computational cost. Third, we investigate a long-standing
evaluation problem: data splitting strategies in molecule retrieval implicitly
trade off data leakage against domain shift. We formalize this tension by
introducing a quantitative measure of distribution shift, and use it to evaluate
splitting strategies in existing benchmarks. All datasets, splits, candidate
sets, and a unified implementation of MSAlign and baselines are publicly
released to support reproducible research.
\end{abstract}
%While the recent release of benchmark datasets such as MassSpecGym and Spectraverse has considerably accelerated the development of novel machine learning approaches, the complexity of data preprocessing pipelines and the lack of unified implementations make methods and results difficult to reproduce and compar
\section{Introduction}

Metabolomics, the study of small molecules in biological systems, is central to understanding disease mechanisms, discovering biomarkers, and developing therapeutics \citep{Wishart_2019_MetabolomicsInvestigatingPhysiological,Quinn_2020_GlobalChemicalEffects}.
To identify small molecules (metabolites) present in biological samples, biochemists rely mainly on Liquid Chromatography coupled with tandem Mass Spectrometry (LC-MS/MS) \citep{Alseekh_2021_MassSpectrometrybasedMetabolomics}. In a nutshell, the different types of ions generated from the molecule entering the mass spectrometer can be selected individually according to their mass-to-charge ratio (m/z) and fragmented at specific collision energies \citep{Domingo-Almenara_2018_AnnotationComputationalSolution}. The resulting measured MS/MS spectrum consists of a list of tuples (m/z, intensity) corresponding to the fragment ions, which provides information about the chemical bonds and substructures of the compound \citep{DeVijlder_2018_TutorialSmallMolecule}.
Recovering the full 2D structure from this spectrum, i.e., \textit{metabolite identification}, is critical yet fundamentally hard \citep{Vaniya_2022_RevisitingCASMICompound, ElAbiead_2025_DiscoveryMetabolitesPrevails, Xu_2025_UnveilingDarkMatter}. The dominant practical approach to this problem relies on spectral library matching, where an experimental spectrum is compared against a database of reference spectra from known compounds \citep{Schymanski_2014_IdentifyingSmallMolecules, Kind_2018_IdentificationSmallMolecules}. However, these libraries cover less than 1\% of the 110 million compounds in PubMed \citep{Bittremieux_2022_CriticalRoleThat}, which leaves most molecules unidentified.

Machine learning offers a way around this coverage bottleneck~\citep{heinonen2012metabolite,Nguyen_2019_RecentAdvancesProspects}, with rapid progress driven by large-scale datasets and the emergence of foundation models dedicated to mass spectrometry~\citep{dreams}. Because the space of metabolites is largely unknown, the most general setting, \textit{de novo} structure elucidation, remains very difficult and its performance is still limited~\citep{Russo_2024_MachineLearningMethods, litsa2021spec2mol, stravs2022msnovelist, wang2025madgen, Bohde_2025_DiffMSDiffusionGeneration}.
We focus instead on the more tractable \textit{molecule retrieval} task. Given a spectrum and a set of candidate molecules (e.g., from PubChem), the goal is to retrieve the correct structure. This mirrors practical settings, where
spectra are searched against large molecular datasets. Most of the proposed ML methods, even if not presented as such, fit the general energy-based framework \citep{lecun2006tutorial}, well suited to tackle structured output prediction as a retrieval task. The approaches to learning the scoring function are divided into two groups: some rely on fingerprint regression \citep{heinonen2012metabolite, Durkhop-2015} or more sophisticated variants \citep{brouard2016fast,el2024sketch,de2026small}, 
while others focus on learning spectra or molecule representations in a contrastive way \citep{kalia2025jestr,chen2026flare,goldman2023mist}. 
In this work, we build on the latter approaches, exploiting the recent advances of multimodal alignment and chemical foundational models \citep{dreams,chemberta}.

\begin{figure}[t]
    \centering
    \includegraphics[width=\linewidth, trim={.7cm .4cm .7cm .3cm}, clip]{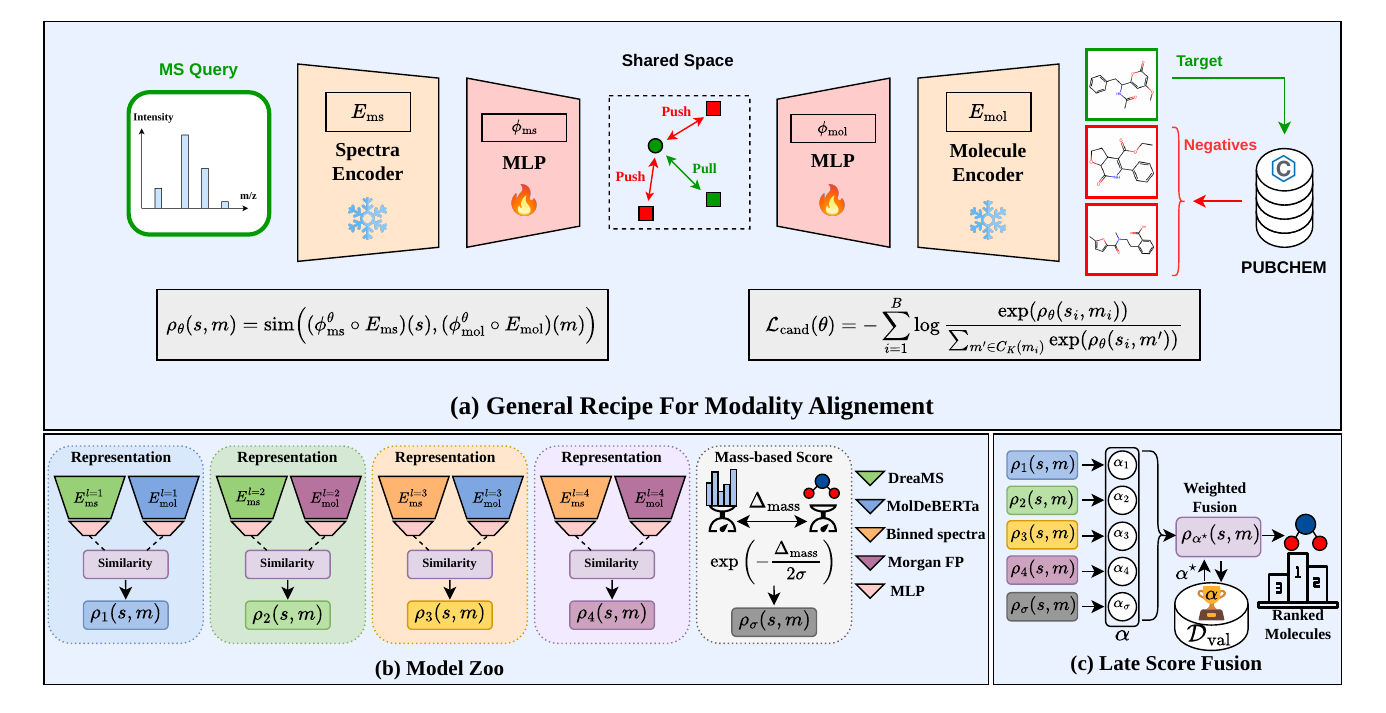}
    \vspace{-0.2cm}
    \caption{(a) Modality-alignment recipe; MSAlign instantiates it with frozen foundation encoders (DreaMS, MolDeBERTa). (b) The recipe yields a zoo of models, complemented by a target-mass score. (c) Late fusion aggregates them via an optimal convex combination of scores.}
    \label{fig:main}
    \vspace{-4mm}
\end{figure}

\paragraph{Contributions.} This paper addresses molecular retrieval from mass spectra through several
complementary contributions detailed below:
\begin{itemize}[leftmargin=10pt, itemsep=0.2em, topsep=0pt, parsep=0pt]
    \item \textbf{A Unified framework} for molecule retrieval, clarifying how prior methods relate (Section~\ref{sec:framework}).
    \item \textbf{MSAlign}, a model that aligns frozen foundation models into a shared spectrum and molecule space, achieving state-of-the-art performance despite lightspeed training. We show that its components are optimal among the alternatives identified in the unified framework (Section~\ref{sec:method}).
    \item \textbf{\msalignplus}, which further improves MSAlign with a score aggregation scheme, combining both alternative representations and one-dimensional chemical information.
    \item \textbf{Distribution shift analysis.} We show that splits that minimize data leakage tend to maximize domain shift, and vice versa. We then introduce a metric to navigate this trade-off and make benchmarks easier to interpret (Section~\ref{sec:split}).
    \item \textbf{Experiments.} We run extensive experiments and ablations that establish state-of-the-art results and validate our design choices (Sections~\ref{subsec:performances}, \ref{subsec:ablation}). 
\end{itemize}
The open-source code for reproducible implementation of MSAlign baselines is provided at:
\begin{center}
\url{https://github.com/KrzakalaPaul/MSAlign-NeurIPS2026}\;\faGithub
\end{center}

\section{A unified framework for molecule retrieval \label{sec:framework}}

In this section, we introduce a unified framework to analyze the different approaches proposed in the recent literature for molecule retrieval from MS/MS spectra. 

\paragraph{Molecule retrieval problem.}
Let $\mathcal{S}$ denote the space of mass spectra and $\mathcal{M}$ the space of molecules. Given a dataset of paired observations $(s_i, m_i)_{i=1}^N \in \bmS \times \bmM$, the goal of Metabolite Identification is to learn a model $f: \bmS\to \bmM$ that predicts the 2D molecular structure associated with a given spectrum.
This problem presents difficult challenges due to the discreteness and combinatorial nature of the output space $\bmM$ and fits the general framework of Structured Output Prediction \citep{bakir2007predicting, nowozin2011structured}. Molecule Identification is generally cast into a Molecule Retrieval problem, where the search is limited to a candidate set $C(s) \subset \bmM$. In that case, $f_\theta$, the retrieval function, based on a scoring function $\rho_\theta: \bmS \times \bmM \to \mathbb{R}$ of parameter $\theta$, writes as follows:
\begin{equation}
f_\theta(s) = \argmax_{m' \in C(s)} \rho_\theta(s, m').
\end{equation}
Then, the goal is to learn a scoring function that assign higher scores to the correct pairs than to the incorrect ones which
connects to the large body of works in energy-based models \citep{lecun2006tutorial}.\looseness=-1

\begin{table*}[t]
\centering
\caption{%
  Comparison of baselines methods within the proposed unified framework.}
\vspace{-0.3cm}
\label{tab:method_comparison}
\setlength{\tabcolsep}{5pt}
\renewcommand{\arraystretch}{1.25}
\small
\resizebox{\columnwidth}{!}{%
\setlength{\tabcolsep}{3pt}
\begin{tabular}{@{}l  cc  cc  cc  cc c@{}}
\toprule

% ── Category headers ────────────────────────────────────────────────────────
&

\multicolumn{2}{c}{\textbf{Spectra encoder}}  &
\multicolumn{2}{c}{\textbf{Molecule encoder}} &
\multicolumn{2}{c}{\textbf{Training objective}} &
\multicolumn{2}{c}{\textbf{Unsupervised data}} &
\multirow{ 2}{*}{\makecell{\textbf{Require} \\ \textbf{Formula}}}
\\

\cmidrule(lr){2-3}
\cmidrule(lr){4-5}
\cmidrule(lr){6-7}
\cmidrule(lr){8-9}

% ── Column headers ───────────────────────────────────────────────────────────
%\phi_{ms}^{\theta_1}\circ E_{ms}

\textbf{Method}
  & \makecell{\textbf{Fixed} ($E_{ms}$)}
  & \makecell{\textbf{Trained} ($\phi_{ms}$)}
   & \makecell{\textbf{Fixed} ($E_{mol}$)}
  & \makecell{\textbf{Trained} ($\phi_{mol}$)}
  & \makecell{\textbf{Pretrain}}
  & \makecell{\textbf{Finetune}}
  & \makecell{\textbf{Spectra}}
  & \makecell{\textbf{Mol.}}
  & 
  \\
  
\midrule

% ── Baselines ────────────────────────────────────────────────────────────────
%\multicolumn{9}{@{}l}{\textit{Baselines}} \\[2pt]

FFN
  & Binarize Spectra & ---
  & Fingerprint  & ---
  & $\mathcal{L}_\text{FP}$ \eqref{eq:regression} & ---
  & $\times$ & $\times$
  & No
  \\

DeepSets
  & Fourier Features  & DeepSets
  & Fingerprint  & ---
  & $\mathcal{L}_\text{FP}$ \eqref{eq:regression} & ---
  & $\times$ & $\times$
  & No
   \\

MIST
  & Peaks Subformulas & Transformer
  & Fingerprint  & MLP
  & $\mathcal{L}_\text{FP}$ \eqref{eq:regression} 
  & $\mathcal{L}_\text{cand}$ \eqref{eq:infoce_candidates}
  & Simulated & Negatives
  & Yes
 \\

Emb-Cos
  & Binarize Spectra  & MLP
  & Fingerprint  & MLP
  & $\mathcal{L}_\text{cand}$ \eqref{eq:infoce_candidates}  & ---
  & $\times$ & Negatives
  & No
 \\

JESTR
  & Binarize Spectra  & MLP
  & Graph& GNN
  & $\mathcal{L}_\text{batch}$ \eqref{eq:infoce_batch}
  & $\mathcal{L}_\text{cand}$ \eqref{eq:infoce_candidates}
  & $\times$ & Negatives
  & No
   \\

FLARE
  & Peaks Subformulas & Transformer
  & Graph& GNN
  & $\mathcal{L}_\text{batch}$ \eqref{eq:infoce_batch}
  & $\mathcal{L}_\text{cand}$ \eqref{eq:infoce_candidates}
  & $\times$ & Negatives
  & Yes
   \\
   
% SAIL
%   & DreaMS & MLP
%   & ChemBERTa & MLP
%   & $\mathcal{L}_\text{batch}$ \eqref{eq:infoce_batch} & ---
%   & DreaMS  & ChemBERTa
%   & No
%  \\

\midrule

% ── Proposed ─────────────────────────────────────────────────────────────────
%\multicolumn{9}{@{}l}{\textit{Proposed}} \\[2pt]

\rowcolor{oursrow}
MSAlign
  & DreaMS & MLP
  & ChemBERTa & MLP
  & $\mathcal{L}_\text{cand}$ \eqref{eq:infoce_candidates}  & ---
  & DreaMS  & ChemBERTa
  & No
 \\
 
\bottomrule
\end{tabular}
}
\vspace{-2mm}
\end{table*}

In our problem, the candidate sets are typically defined via a mass filter, i.e., $C(s) = \{m' \in \bmM : \mathrm{abs}(|m'| - |m_s|) < \epsilon\}$ where $|m'|$ (resp. $|m_s|$) is the mass of a candidate (resp. target) molecule and $\epsilon$ a threshold. This is a reasonable assumption, as the determination of the mass of the neutral molecule given the MS/MS spectrum (i.e., the determination of the ion species that was fragmented) is a well-managed task both manually and automatically \citep{Vaniya_2022_RevisitingCASMICompound}.\looseness=-1

\paragraph{Scoring function and training loss.}
We focus here on the large class of scoring functions $\rho_\theta$ defined as a similarity {\it sim}: $\bmZ \times \bmZ \to \mathbb{R}^+$  over a shared representation $\bmZ$ of spectra and molecules, i.e. \looseness=-1
\begin{equation}
\rho_\theta(s, m) = \mathrm{sim}(g_\theta(s), h_\theta(m))
\end{equation}
where $g_\theta: \mathcal{S} \to \bmZ$ and $h_\theta: \mathcal{M} \to \bmZ$ are encoders or {\it feature maps} and $\mathrm{sim}$ is typically the cosine similarity. The two following approaches have been proposed to learn the score:

\textit{Retrieval through regression.} A first approach is to concentrate on learning the feature map $g_{\theta}$ while keeping $h_{\theta}(m)=h(m)$ fixed, either as a molecular fingerprint $FP(m)$ i.e. a binary vector representing the presence/absence of certain substructures \citep{heinonen2012metabolite, goldman2023mist, ji2020predicting} or as a feature vector in a Hilbert space \citep{brouard2016fast,brogat2022learning}. Learning the score function boils down to estimate the regression of the variable $h(m)$ by the variable $s$. 
Taking the example of fingerprint regression, the loss function can then be chosen as:\looseness=-1
\begin{equation}
\mathcal{L}_\text{FP}(\theta) = - \frac{1}{B} \sum_{i=1}^B \mathrm{sim}(g_\theta(s_i), \text{FP}(m_i)).
\label{eq:regression}
\end{equation}

\textit{Retrieval through constrastive learning.} The alternative is to jointly learn the feature maps $g_\theta$ and $h_\theta$. This formulation connects self-supervised multimodal alignment~\citep{radford2021CLIP} with supervised retrieval, naturally leading to contrastive learning objectives~\citep{chen2020simple,wang2020contrastive}. The most common choice is the InfoNCE loss~\citep{kalia2025jestr,chen2026flare,de2026small}, which encourages the similarity of the true pair $(s, m)$ to be higher than that of negative pairs $(s, m')$ and practical implementation differ in the definition of negatives pairs. As training is performed through mini-batch iteration, a common choice is to use the other molecules in the batch as negatives, leading to the following \textit{in-batch} InfoNCE loss:
\begin{equation}
\mathcal{L}_\text{batch}(\theta) = - \frac{1}{B} \sum_{i=1}^B \log \frac{\exp(\rho_\theta(s_i, m_i)/\epsilon)}{\sum_{j=1}^B\exp(\rho_\theta(s_i, m_j)/\epsilon)}.
\label{eq:infoce_batch}
\end{equation}
However, this approach is limited by the fact that in-batch negatives are not necessarily hard to distinguish, making the training loss less informative. A large body of work attempt to address this limitation through hard negative mining or by scaling the number of negatives~\citep{robinson2020contrastive, thirukovalluru2025breaking, zhang2025sail}. In the next section, we argue that these strategies are unnecessary for molecular retrieval, as the chemically defined candidate set $C(s_i)$ provides a natural source of hard negatives which yields the \textit{candidate-based} InfoNCE loss \eqref{eq:infoce_candidates}.

\paragraph{Architecture and data representation.}
The second key design choice is the architecture of the encoders $g_\theta$ and $h_\theta$ and the underlying data representation for each modality \citep{pollmann2026bridging}.\looseness=-1

\textit{For spectra}, common approaches include discretizing the m/z axis into fixed bins~\citep{kalia2025jestr,de2026small}, applying convolutional neural networks~\citep{kudriavtseva2021deep}, or modeling spectra as sequences of peaks (optionally enriched with formula annotations) and processed by transformer architectures~\citep{goldman2023mist,chen2026flare, deJonge_2025_BinNotBin}.
%\ET{[Pour information, une référence compare des représentations de spectres sous forme de bin, d'ensemble de pics, ou de suites de pics avec des graphes \citep{deJonge_2025_BinNotBin}.]}

\textit{For molecules}, classical representations rely on handcrafted fingerprints, optionally refined with an MLP~\citep{goldman2023mist}, while more recent methods use graph neural networks (GNNs) to directly encode molecular structures~\citep{kalia2025jestr,chen2026flare}. An alternative paradigm represents molecules as SMILES or SELFIES strings and leverages transformer-based language models to learn molecular embeddings~\citep{elser2023mass2smiles, han2025ms}. 

\paragraph{Unified framework.}
We summarize the possible encoding choices in a compact formulation:
\begin{align}
    g_{\theta_1}(s) = (\phi_{ms}^{\theta_1}\circ E_{ms})(s), \quad
   h_{\theta_2}(m) = (\phi_{mol}^{\theta_2}\circ E_{mol})(m),
\end{align}
in which $E_{ms}$ and $E_{mol}$ are non-trainable preprocessing steps (e.g. binning for spectra, fingerprinting for molecules) and $\phi_{ms}^{\theta_1}$ and $\phi_{mol}^{\theta_2}$ are trainable encoders (e.g. MLP, GNN, transformer). Under this formulation, a large body of work can be reduced to the choice of $E_{ms}$, $E_{mol}$, $\phi_{ms}^{\theta_1}$ and $\phi_{mol}^{\theta_2}$, as well as the loss function among those presented in \eqref{eq:regression}, \eqref{eq:infoce_batch} or \eqref{eq:infoce_candidates}. Table \ref{tab:method_comparison} summarizes the comparison between recent methods along with the key design features discussed above.

\section{MSAlign: lightweight model with state-of-the-art performances }\label{sec:method}

\paragraph{Foundation model alignment.} We propose a lightweight implementation of the general framework introduced above where $E_{ms}:\mathcal{S}\mapsto \mathbb{R}^{d_\mathrm{ms}}$, $E_{mol}:\mathcal{M}\mapsto \mathbb{R}^{d_\mathrm{mol}}$ are fixed vector representation for each modality, and $\phi_{ms}^{\theta_1}:\mathbb{R}^{d_\mathrm{ms}}\mapsto \mathbb{R}^d$, $\phi_{mol}^{\theta_2}:\mathbb{R}^{d_\mathrm{mol}}\mapsto \mathbb{R}^d$ are lightweight MLP adapters. Inspired by recent works in vision--language alignment \citep{zhang2025sail,roschmann2026sotalign} and the Platonic Representation Hypothesis~\citep{huh2024platonic} we propose to align frozen foundation-model representations and leverage DreaMS~\citep{dreams} as $E_\mathrm{ms}$ and MolDeBERTa~\citep{de2026moldeberta} as $E_\mathrm{mol}$. Critically, since the foundation-model are frozen, the representations $E_{ms}(s)$ and $E_{mol}(m)$ can be precomputed which enables lightspeed training and
makes it possible to leverage the rich candidate-based InfoNCE loss described below. We refer to Appendix~\ref{appendix:details} for additional architecture details and Table~\ref{tab:representations} for comparison with alternative frozen encoders.\looseness=-1

\paragraph{Candidate-based training.} We argue that the \textit{in-batch} InfoNCE loss \eqref{eq:infoce_batch} is not suited for molecular retrieval, as molecules in the same batch are typically easy to distinguish (by mass or formula for instance). We refer to Appendix~\ref{appendix:infonce} for further discussion on the topic. This is a classical limitation in contrastive learning, which typically calls for hard negative sampling strategies \citep{robinson2020contrastive}. Fortunately, in our setting, the chemically defined candidate set $C(s_i)$ provides a natural source of hard negatives which can be exploited via the \textit{candidate-based} InfoNCE loss:
\begin{equation}
\mathcal{L}_\text{cand}(\theta) = - \frac{1}{B} \sum_{i=1}^B \log \frac{\exp(\rho_\theta(s_i, m_i)/\epsilon)}{\sum_{m' \in C_K(s_i)} \exp(\rho_\theta(s_i, m')/\epsilon)},
\label{eq:infoce_candidates}
\end{equation}
where $C_K(s_i) \subset C(s_i)$ is a subset of $K$ negatives sampled from the candidate set. This loss is much more informative than its \textit{in-batch} counterpart, and in particular it is a valid convex surrogate of the top-1 error as formalized in Proposition~\ref{prop:msalign_top1}. The only limitation of \eqref{eq:infoce_candidates} is its computational costs as it requires $K \times B$ similarity evaluations $\rho_\theta(s_i, m')$ with limited reuse of molecule embeddings across spectra. Critically, the pre-computation of embeddings, make the use of this loss tractable and even surprisingly fast, with less than 2minutes/epochs on MassSpecGym as reported in Table~\ref{tab:train_time}. 

\paragraph{Late score fusion.} The lightweight recipe described above makes it inexpensive to train several alignment models by varying the spectral and molecular frozen representations, yielding a model zoo (Figure~\ref{fig:main}b). We hypothesize that models based on different representations yields uncorrelated errors (this is verified experimentally, see Figure~\ref{fig:losses_figure}c), which naturally calls for a model aggregation strategy.

Following earlier work on molecular identification~\citep{ruttkies2016metfrag,ruttkies2019improving,Brouard_2019_ImprovedSmallMolecule,li2024ensemble}, we combine $L$ fixed scores through a convex combination:
\begin{equation}
    \rho_{\boldsymbol\alpha}(s,m)
    =\sum_{\ell=1}^{L}\alpha_\ell\rho_\ell(s,m),
    \qquad
   \boldsymbol\alpha^\star
    \in\argmax_{\boldsymbol\alpha\in\Delta^{L-1}} \
    \mathrm{R@1}_{\mathrm{val}}(\boldsymbol\alpha).
\label{eq:msalign_fusion}
\end{equation}
where $\mathrm{R@1}_{\mathrm{val}}$ is the top-1 accuracy on the held-out validation split and $\Delta^{L-1}$ is the simplex. This is a non-convex problem in $L$ weights which we approximate using a smooth surrogate and validation-based temperature selection. The procedure is detailed in Appendix~\ref{appendix:fusion_algorithm} and Algorithm~\ref{alg:fusion_calibration}.

Fusion also accommodates one-dimensional chemical information. Let $\widehat M(s)$ be the neutral mass estimated from the precursor $m/z$, accounting for charge and adduct, and $M(m)$ the candidate's monoisotopic mass. We define
\begin{equation}
    e_{\mathrm{ppm}}(s,m)
    =10^6\frac{\widehat M(s)-M(m)}{M(m)},
    \qquad
    \rho_{\sigma}^{\mathrm{mass}}(s,m)
    =\exp\!\left[-\frac{e_{\mathrm{ppm}}(s,m)^2}{2\sigma^2}\right].
\label{eq:msalign_mass_score}
\end{equation}

Ultimately, the proposed model,
\textbf{\msalignplus}, combines $L=7$ scores: four alignment scores from pairing binned spectra or DreaMS with Morgan fingerprints or MolDeBERTa, and three mass scores with bandwidths $\sigma\in\{0.01,0.1,1\}$ ppm. Table~\ref{tab:scorers} isolates the gains from alignment-score fusion and the additional mass scores, while Table~\ref{tab:main-results} compares the resulting model with competing methods.

\textbf{MSAlign} refers to the raw model achieved by aligning DreaMS and MolDeBERTa representations (no score fusion), it is the strongest individual components of \textbf{\msalignplus} (Table~\ref{tab:scorers}).

\section{Assessing the challenges of data splitting in metabolite \label{sec:split}}

Before moving to the experimental evaluation of MSAlign, we first discuss a critical aspect of the experimental protocol: the choice of data splitting strategy.

\begin{wrapfigure}{r}{0.38\linewidth}
    \vspace{-0.2cm}
    \centering
    \includegraphics[width=\linewidth]{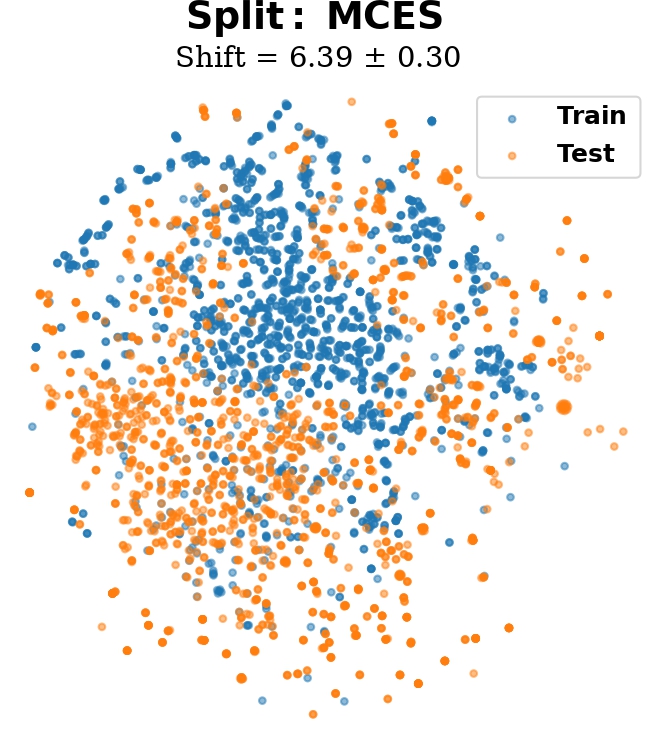}
    \vspace{-0.4cm}
    \caption{t-SNE visualization of the MassSpecGym MCES splits.}
    \label{fig:tsne_shift}
    \vspace{-0.5cm}
\end{wrapfigure}
\paragraph{Challenge.}
Data splitting is a non-trivial aspect of MS/MS benchmarking. The same molecule can appear in multiple spectra acquired under different experimental conditions, making naive random splits prone to \textbf{data leakage}. To mitigate this issue, existing benchmarks typically enforce some degree of molecular dissimilarity across folds. However, increasingly stringent splitting criteria can induce substantial \textbf{distribution shifts} between training and test sets~\citep{rakhshaninejad2026reliable}, which may not reflect practical deployment settings and may require dedicated domain adaptation techniques~\citep{farahani2021brief}.

A wide range of splitting strategies have therefore been proposed, differing primarily in how strongly they separate molecules across folds. For instance, the MassSpecGym MCES split requires molecules from different folds to have an edit distance of at least $10$~\citep{massspecgym}, whereas NPLIB1 only requires the corresponding structures to be distinct (InChIKey split). We refer to Appendix~\ref{appendix:splits} for a detailed presentation of alternative criterions.

Yet, to the best of our knowledge, there is no principled way to compare these strategies or quantify the distribution shift they induce. This makes it difficult to assess the trade-off between preventing data leakage and preserving a realistic train--test distribution. To address this issue, we introduce a quantitative metric that measures the degree of distribution shift induced by a given splitting strategy.

\paragraph{Quantitative measure.} We propose to compute the Sliced Wasserstein distance \citep{bonneel2015sliced} between the training and test distributions. This requires encoding each sample $x = (s, m)$ as a vector
\begin{equation}
   F(x) = \operatorname{concat}\left(\frac{\operatorname{encode}_s(s)}{\|\operatorname{encode}_s(s)\|_2}, \frac{\operatorname{encode}_m(m)}{\|\operatorname{encode}_m(m)\|_2}\right) \in \mathbb{R}^{d_s + d_m},
\label{eq:split_feature_map}
\end{equation}
and we report the results both for foundation-model embeddings ($\operatorname{encode}_s = \mathrm{DreaMS}$ and $\operatorname{encode}_m = \mathrm{MolDeBERTa}$) and for simple vectorization baselines i.e. a fingerprint ($\operatorname{encode}_m = \mathrm{Morgan}$) for molecules and a binned spectrum ($\operatorname{encode}_s = \mathrm{Binned}$) for spectra.  

Finally, to make the metric more interpretable, we normalize it by the value obtained with a random split. Formally, let $\mathcal{D}_\text{train}$ and $\mathcal{D}_\text{test}$ denote the training and test sets to evaluate, and let $(\mathcal{D}_1, \mathcal{D}_2)$ be a random split such that $\mathcal{D}= \mathcal{D}_1 \cup \mathcal{D}_2 = \mathcal{D}_\text{train} \cup \mathcal{D}_\text{test}$. We define the normalized sliced Wasserstein distance by
\begin{equation}
    \mathrm{Shift}(\mathcal{D}_\text{train}, \mathcal{D}_\text{test}) = \frac{\mathrm{SWD}(\mathcal{D}_\text{train}, \mathcal{D}_\text{test})}{\mathbb{E}_{(\mathcal{D}_1, \mathcal{D}_2)}[\mathrm{SWD}(\mathcal{D}_1, \mathcal{D}_2)]}.
\label{eq:normalized_shift}
\end{equation}
where $\mathrm{SWD}$ is the Sliced Wasserstein distance between the empirical distributions of the two sets. We use the implementation from the POT library \citep{flamary2024pot} with 100 random projections.

\begin{table}[t]
\centering
\small
\resizebox{0.9\columnwidth}{!}{%
\begin{tabular}{@{}lcccccc@{}}
\toprule
& \multicolumn{6}{c}{\textbf{Data Splitting Criterion}}\\
\cmidrule{2-7}
\textbf{Encoders} & \textbf{Random} & \textbf{InChIKey} & \textbf{Formula} & \textbf{Murcko Scaffold} & \textbf{MCES} & \textbf{Murcko Histogram} \\
\midrule
$\mathrm{Binned}-\mathrm{Morgan}$ & $1.00 \pm 0.03$ & $4.71 \pm 0.17$ & $5.19 \pm 0.35$ & $9.24 \pm 1.02$ & $9.10 \pm 0.33$ & $15.22 \pm 4.83$ \\
$\mathrm{DreaMS}-\mathrm{Morgan}$ & $1.00 \pm 0.03$ & $4.99 \pm 0.20$ & $5.71 \pm 0.35$ & $9.76 \pm 1.26$ & $11.95 \pm 0.53$ & $17.66 \pm 5.64$ \\
$\mathrm{Binned}-\mathrm{MolDeBERTa}$ & $1.00 \pm 0.02$ & $4.00 \pm 0.42$ & $4.65 \pm 0.27$ & $7.01 \pm 0.71$ & $12.41 \pm 0.86$ & $15.12 \pm 4.32$ \\
$\mathrm{DreaMS}-\mathrm{MolDeBERTa}$ & $1.00 \pm 0.04$ & $4.44 \pm 0.28$ & $4.98 \pm 0.44$ & $8.51 \pm 0.57$ & $15.89 \pm 0.75$ & $18.42 \pm 5.62$ \\
\bottomrule
\end{tabular}
}
\vspace{+0.2cm}
\caption{Normalized Sliced Wasserstein shift between the training and test sets for the different data-splitting strategies. We report average and standard deviation over five splits, except for MCES which has two splits.}
\label{tab:splits}
\vspace{-0.6cm}
\end{table}

\paragraph{Analysis on MassSpecGym.} In Table~\ref{tab:splits}, we report the proposed metric on MassSpecGym across the different splitting strategies and encoders $\operatorname{encode}_m$, $\operatorname{encode}_s$. We observe that the overall ranking is robust to the choice of encoders. MCES and Murcko splits are consistently the splits that induce the largest shifts.  We illustrate this in Figure~\ref{fig:tsne_shift} by plotting a 2D visualization of the joint latent space \eqref{eq:split_feature_map} for the MCES split: we observe that the test set forms a distinct cluster far from the training data. 

Hence, those splits are particularly challenging, and can be used to measure the out-of-distribution generalization capabilities of a model, this may be relevant for specific applications such as de novo drug discovery. However, such a strong shift may not reflect practical deployment in retrieval settings, where test molecules are likely to share some similarity with those in the training set. We encourage future works to address these challenging splits with dedicated domain adaptation techniques.

Overall, the Formula and 2D InChIKey splits provide a more balanced trade-off, with Shift values around 4-5. Of the two, we recommend Formula-split as it is strictly more challenging (two molecules with the same InChiKey necessarily have the same formula), and we adopt it for all subsequent experiments. For completeness, MassSpecGym's harder original MCES split is also evaluated.

\section{Experimental Results}

In this section, we first present the experimental setting \ref{subsec:setting}, then we report the retrieval performances of MSAlign and \msalignplus compared to a range of baselines \ref{subsec:performances}. Finally, we validate step by step the key design choices of our method \ref{subsec:ablation}.

\subsection{Experimental Setting \label{subsec:setting}}

\paragraph{Supervised Datasets}
We consider two open-source benchmarks.
\textbf{MassSpecGym}~\citep{massspecgym} provides 231,104 carefully curated and
standardized pairs, and is rapidly becoming the community standard for benchmarking.
\textbf{Spectraverse}~\citep{spectraverse} is the larger (488,797 pairs) and
provides broader coverage of adducts and ionization modes, with more 50\% of spectra
corresponding to non-\mbox{[M+H]$^+$} adducts. 
Note that pair counts overestimate chemical diversity as molecules frequently appear
under multiple ionization conditions; Table~\ref{tab:dataset_stats} reports unique
molecule counts along with pair counts for all datasets.

\paragraph{Candidates Sets.} Following the standardized protocol of MassSpecGym~\citep{massspecgym}, candidate
molecules are retrieved from PubChem~\citep{pubchem} (118M compounds) by
selecting structures whose molecular mass matches the target molecule within a 10\, ppm tolerance, prioritizing compounds sampled from databases of decreasing metabolite relevance (see Algorithm~1
of~\cite{massspecgym} for details).
We keep $K=256$ candidates per spectrum; as illustrated in Figure~\ref{fig:losses_figure}, these
candidates are typically structurally close to the target, making retrieval
challenging even at this moderate pool size. 

\paragraph{Data Splitting and seeds.} Following the discussion in Section~\ref{sec:split}, we split the dataset using the chemical formula as a criterion, ensuring that molecules in the train/validation/test sets have distinct formulas, and report results averaged over three random seeds of the splitting algorithm. For completeness, we also report results on the official MassSpecGym MCES split. Since generating alternative MCES splits is computationally expensive, we instead average over its two variants obtained by swapping the validation and test sets, an important source of variability since the validation set is used for both calibration and checkpoint selection by MSAlign and baselines.

\paragraph{Baselines.}
We propose a comprehensive list
of recent baselines: \textbf{
DeepSets} is a fingerprint prediction baseline introduced in MassSpecGym
\citep{massspecgym}. To maximize the performances, we use the Fourier Features variants and increase the number of
epochs to 50 which significantly improves this baseline compared to default settings.
\textbf{MIST} \citep{goldman2023mist} is more complex model that pretrains with
fingerprint prediction, finetunes with contrastive learning, and augments the
data with a forward model {for spectra simulation}. \textbf{JESTR} \citep{kalia2025jestr} is a
simple contrastive model that pretrains with batch InfoNCE \eqref{eq:infoce_batch}
and finetunes with the more computationally expensive candidates InfoNCE \eqref{eq:infoce_candidates}. \textbf{MVP} \citep{zhou2026learning} builds on JESTR by introducing a multi-view setting (fingerprint, graph, spectra and consensus spectra).
\textbf{FLARE} \citep{chen2026flare} is a direct follow-up work that considers a stronger spectra
encoder leveraging peak annotation {with subformulas} and a more
interpretable, richer similarity. 
\textbf{Emb-Cos} is the
strongest of the four models proposed in \citep{de2026small}; similarly to
MSAlign, it uses candidates InfoNCE to learn a shared space for molecules and
mass spectra, but it does not use pretrained encoders or score fusion. 

%\paragraph{Reproductibility.}
%For fair and reproducible results we re-implement all methods within our open source code base. The only exception to this is MIST which is a complex framework that proves challenging to re-implement from scratch  \citep{heirman2024reusability}.

\begin{table*}[t]
\centering
\caption{Recall score across datasets and models. We report the standard deviation \textbf{over 3 random splitting} of the dataset. Methods reported on the right require an oracle to predict the target's formula.}
\vspace{-0.1cm}
\label{tab:main-results}
\resizebox{\textwidth}{!}{%
\begin{tabular}{llccccc|cccc}
\toprule
 &  & \multicolumn{5}{c|}{Without molecular formula} & \multicolumn{4}{c}{Requires target molecular formula} \\
\cmidrule(lr){3-7}\cmidrule(lr){8-11}
\textbf{Dataset} & \textbf{Metric} & \textbf{DeepSet} & \textbf{JESTR} & \textbf{Emb-Cos} & \textbf{MSAlign} & \textbf{\msalignplus} & \textbf{MIST} & \textbf{MVP} & \textbf{FLARE} & \textbf{\makecell{\msalignplus \\ (+Filter)}} \\
\midrule
\multirow{3}{*}{SpectraVerse} & R@1 & 2.0 {\scriptsize $\pm$ 0.4} & 5.4 {\scriptsize $\pm$ 0.2} & 26.6 {\scriptsize $\pm$ 1.9} & \underline{31.8} {\scriptsize $\pm$ 2.6} & \textbf{41.3} {\scriptsize $\pm$ 2.7} & \underline{22.7} {\scriptsize $\pm$ 1.6} & 9.6 {\scriptsize $\pm$ 0.6} & 14.2 {\scriptsize $\pm$ 1.1} & \textbf{44.4} {\scriptsize $\pm$ 2.6} \\
 & R@5 & 8.4 {\scriptsize $\pm$ 1.0} & 17.5 {\scriptsize $\pm$ 1.3} & 49.5 {\scriptsize $\pm$ 1.0} & \underline{58.2} {\scriptsize $\pm$ 1.9} & \textbf{68.3} {\scriptsize $\pm$ 1.7} & \underline{41.5} {\scriptsize $\pm$ 0.8} & 25.9 {\scriptsize $\pm$ 1.8} & 34.7 {\scriptsize $\pm$ 2.6} & \textbf{71.4} {\scriptsize $\pm$ 1.5} \\
 & R@20 & 22.9 {\scriptsize $\pm$ 0.9} & 39.4 {\scriptsize $\pm$ 2.0} & 71.0 {\scriptsize $\pm$ 1.3} & \underline{78.3} {\scriptsize $\pm$ 1.7} & \textbf{85.4} {\scriptsize $\pm$ 0.9} & \underline{64.1} {\scriptsize $\pm$ 0.5} & 48.3 {\scriptsize $\pm$ 0.7} & 60.3 {\scriptsize $\pm$ 2.0} & \textbf{87.5} {\scriptsize $\pm$ 0.9} \\
\midrule
\multirow{3}{*}{\makecell{MassSpecGym \\ (Formula Split)}} & R@1 & 4.7 {\scriptsize $\pm$ 0.6} & 13.9 {\scriptsize $\pm$ 0.5} & 42.7 {\scriptsize $\pm$ 3.0} & \underline{47.5} {\scriptsize $\pm$ 3.4} & \textbf{58.2} {\scriptsize $\pm$ 3.4} & \underline{41.4} {\scriptsize $\pm$ 1.7} & 23.3 {\scriptsize $\pm$ 3.4} & 32.8 {\scriptsize $\pm$ 2.5} & \textbf{60.4} {\scriptsize $\pm$ 3.7} \\
 & R@5 & 13.0 {\scriptsize $\pm$ 1.3} & 32.5 {\scriptsize $\pm$ 0.8} & 65.2 {\scriptsize $\pm$ 2.8} & \underline{71.3} {\scriptsize $\pm$ 2.6} & \textbf{80.7} {\scriptsize $\pm$ 3.0} & 60.8 {\scriptsize $\pm$ 0.7} & 47.3 {\scriptsize $\pm$ 2.0} & \underline{61.8} {\scriptsize $\pm$ 2.1} & \textbf{82.1} {\scriptsize $\pm$ 3.3} \\
 & R@20 & 27.5 {\scriptsize $\pm$ 2.0} & 57.0 {\scriptsize $\pm$ 0.7} & 80.4 {\scriptsize $\pm$ 1.6} & \underline{87.0} {\scriptsize $\pm$ 1.1} & \textbf{92.8} {\scriptsize $\pm$ 1.1} & 76.2 {\scriptsize $\pm$ 0.9} & 71.8 {\scriptsize $\pm$ 0.6} & \underline{82.3} {\scriptsize $\pm$ 1.5} & \textbf{93.6} {\scriptsize $\pm$ 1.1} \\
\midrule
\multirow{3}{*}{\makecell{MassSpecGym \\ (MCES Split)}} & R@1 & 1.9 {\scriptsize $\pm$ 0.5} & 7.9 {\scriptsize $\pm$ 1.1} & 12.6 {\scriptsize $\pm$ 0.2} & \underline{19.0} {\scriptsize $\pm$ 3.2} & \textbf{30.8} {\scriptsize $\pm$ 2.5} & 16.4 {\scriptsize $\pm$ 0.1} & 25.6 {\scriptsize $\pm$ 0.8} & \underline{28.4} {\scriptsize $\pm$ 2.1} & \textbf{37.7} {\scriptsize $\pm$ 1.3} \\
 & R@5 & 8.4 {\scriptsize $\pm$ 0.6} & 22.3 {\scriptsize $\pm$ 1.4} & 29.9 {\scriptsize $\pm$ 0.2} & \underline{41.2} {\scriptsize $\pm$ 2.5} & \textbf{59.3} {\scriptsize $\pm$ 0.7} & 35.8 {\scriptsize $\pm$ 0.7} & 55.3 {\scriptsize $\pm$ 0.1} & \underline{60.2} {\scriptsize $\pm$ 1.4} & \textbf{67.2} {\scriptsize $\pm$ 1.0} \\
 & R@20 & 21.8 {\scriptsize $\pm$ 1.4} & 44.8 {\scriptsize $\pm$ 1.9} & 51.6 {\scriptsize $\pm$ 0.9} & \underline{64.9} {\scriptsize $\pm$ 0.9} & \textbf{82.0} {\scriptsize $\pm$ 0.8} & 60.2 {\scriptsize $\pm$ 2.3} & 79.9 {\scriptsize $\pm$ 2.3} & \underline{82.8} {\scriptsize $\pm$ 2.8} & \textbf{87.1} {\scriptsize $\pm$ 2.3} \\
\bottomrule
\end{tabular}%
}
\vspace{-0.2cm}
\end{table*}

\subsection{Retrieval Performances and comparison to competitors \label{subsec:performances}}

The performances of the \textbf{MSAlign} and the baselines presented above, are collected in Table~\ref{tab:main-results}. 

\paragraph{Formula oracle.} We note that some baselines require the molecular formula of the target to be provided. This is a strong assumption as in practice it is unknown and must be estimated, which remains an active research problem~\citep{hong2025fiddle, Duhrkop_2019_SIRIUS4Rapid, Xing_2023_BUDDYMolecularFormula}. Consequently, we propose to explicitly distinguish the methods that make this assumption (right panel) from those that don't (left panel). For fair comparison, we propose a simple variant of our method (\textbf{\msalignplus{} + Filter}) that leverage this oracle by filtering the candidates to keep only those matching the ground-truth formula.

\paragraph{Results.} Overall, we observe that \textbf{MSAlign} and its variants achieves strikingly strong results across settings.
Standalone \textbf{MSAlign} already provides a strong and computationally efficient baseline: it achieves the best mean recall on \textbf{Spectraverse} and \textbf{MassSpecGym} formula splits and outperforms all formula agnostic methods on \textbf{MassSpecGym} MCES split, while also being the fastest method to train (Table~\ref{tab:train_time}). Each proposed extension further improves retrieval performance. Score fusion increases R@1 by $9.5$, $10.7$, and $11.8$ percentage points on SpectraVerse, MassSpecGym Formula, and MassSpecGym MCES, respectively. When the ground-truth molecular formula is additionally provided, formula filtering yields another gain of $3.1$, $2.2$, and $6.9$ percentage points. Consequently, the complete oracle variant, \textbf{\msalignplus{} + Filter}, establishes a substantial performance gap over the strongest competing method in each setting, outperforming the second best method by $17.8$, $17.7$ and $9.3$ percent, respectively. Finally, we note some discrepancies with previously reported results: JESTR performs below its original report, which echoes issues discussed in public implementations\footnote{\url{https://github.com/HassounLab/JESTR1/issues/5}}. We observe a similar trend for FLARE, potentially due to shared implementation components.

\subsection{Ablation studies and interpretation of the results \label{subsec:ablation}}

\begin{table}[t]
\centering
\caption{MassSpecGym performance (Formula split) across encoder pairs. The upper grid compares foundational and classical representations; the lower grid compares molecular foundation models.\looseness=-1}
\vspace{0.1cm}
\label{tab:representations}
\footnotesize
%\resizebox{0.75\textwidth}{!}{%
\begin{tabular}{llccc}
\toprule
\multicolumn{2}{c}{\textbf{Encoders}} & \multicolumn{3}{c}{\textbf{MassSpecGym}}\\
\cmidrule(lr){1-2}\cmidrule(lr){3-5}
\textbf{Spectra} (dim) & \textbf{Molecules} (dim) & \textbf{R@1} & \textbf{R@5} & \textbf{R@20} \\
\midrule
%Binned (10050) & Fingerprint (4096) & 34.3 {\scriptsize $\pm$ 1.3} & 56.2 {\scriptsize $\pm$ 2.4} & 71.8 {\scriptsize $\pm$ 1.3} \\
Binned (10050) & Fingerprint (4096) & 34.3 {\scriptsize $\pm$ 1.3} & 56.2 {\scriptsize $\pm$ 2.4} & 71.8 {\scriptsize $\pm$ 1.3} \\
DreaMS (1024) & Fingerprint (4096) & 38.6 {\scriptsize $\pm$ 2.8} & 60.4 {\scriptsize $\pm$ 2.1} & 77.6 {\scriptsize $\pm$ 0.8} \\
Binned (10050) & MolDeBERTa (768) & 44.5 {\scriptsize $\pm$ 5.1} & 68.5 {\scriptsize $\pm$ 3.9} & 84.9 {\scriptsize $\pm$ 0.7} \\
\rowcolor{oursrow}
DreaMS (1024) & MolDeBERTa (768) & \textbf{47.5} {\scriptsize $\pm$ 3.4} & \underline{71.3} {\scriptsize $\pm$ 2.6} & \underline{87.0} {\scriptsize $\pm$ 1.1} \\
\midrule
DreaMS (1024) & ChemBERTa-13M (768) & 46.9 {\scriptsize $\pm$ 3.0} & \textbf{71.6} {\scriptsize $\pm$ 1.7} & \textbf{87.8} {\scriptsize $\pm$ 1.6} \\
DreaMS (1024) & IBM MoLFormer (768) & 46.5 {\scriptsize $\pm$ 5.6} & 69.1 {\scriptsize $\pm$ 3.1} & 85.6 {\scriptsize $\pm$ 0.9} \\
DreaMS (1024) & GROVER (3200) & 42.2 {\scriptsize $\pm$ 5.0} & 67.4 {\scriptsize $\pm$ 2.6} & 84.9 {\scriptsize $\pm$ 0.7} \\
DreaMS (1024) & MHG-GED (1024) & \underline{47.0} {\scriptsize $\pm$ 5.2} & 70.3 {\scriptsize $\pm$ 2.1} & 85.9 {\scriptsize $\pm$ 1.8} \\

\bottomrule
\end{tabular}
%}
\end{table}

\paragraph{Encoders.} In Section~\ref{sec:method}, we introduced the alignment of precomputed representations as a general recipe for molecular retrieval. MSAlign instantiates this approach by aligning DreaMS and MolDeBERTa representations. We now assess this default choice against a broad range of alternative encoders. On the molecular side, we consider representations ranging from classical Morgan fingerprints~\citep{ucak2023reconstruction} to SMILES-based~\citep{chemberta, de2026moldeberta, ross2022large} and graph-based foundation models~\citep{grover,mhg}. For spectra, we compare DreaMS~\citep{dreams} with a classical histogram representation using bins of width $0.1$Da~\citep{kalia2025jestr}.

\begin{figure}[t]
    \centering
\includegraphics[width=\linewidth, trim={1.4cm 0.5cm 1.5cm .7cm}, clip]{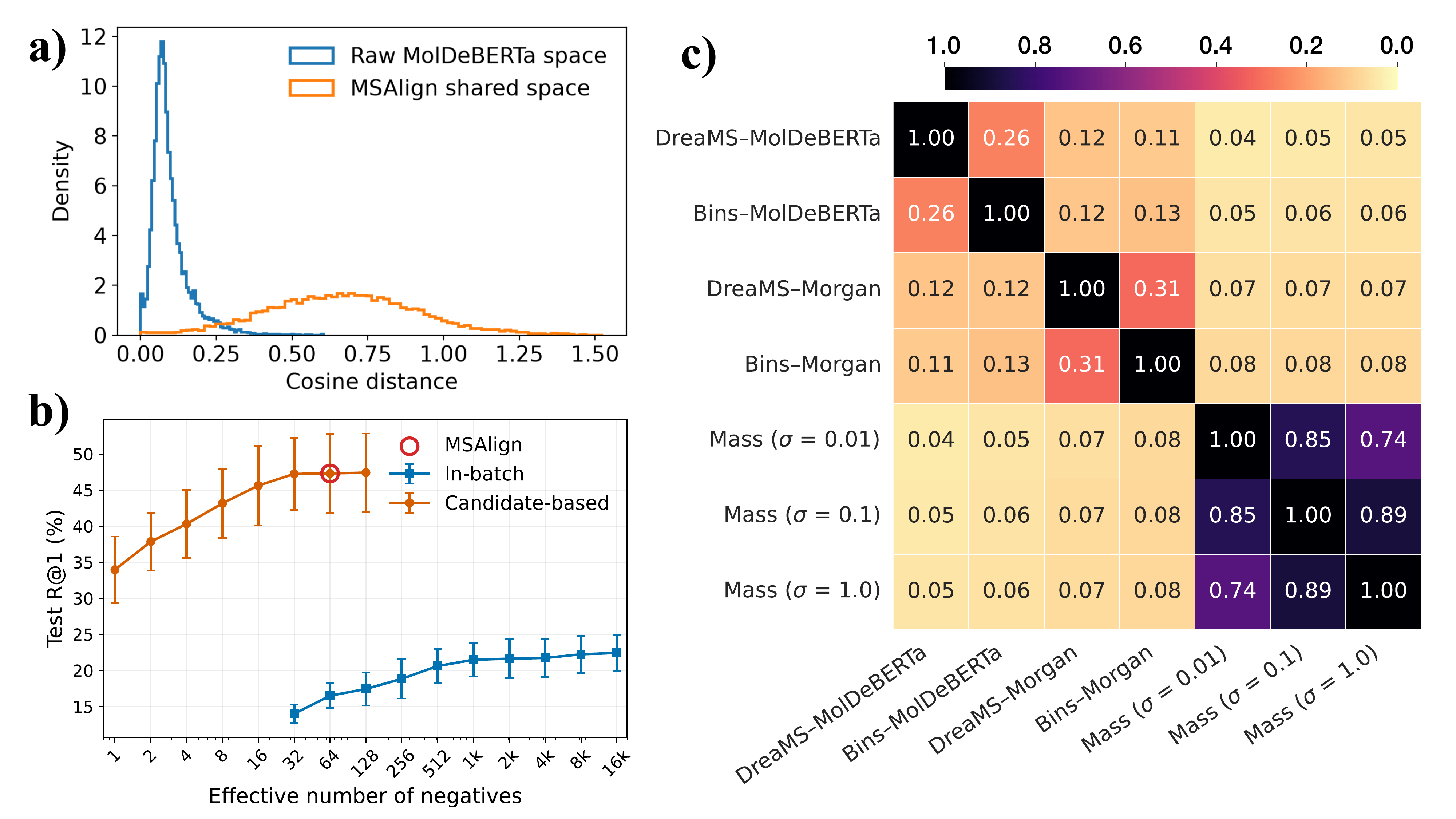}
\vspace{-0.3cm}
    \caption{\textbf{a)} distribution of pairwise cosine distance between pairs of candidates $(m',m'') \in C(s)$. Candidates are very close in MolDeBERTa space ($E[d(m',m'')]=0.09$) and thus hard to distinguish. On the contrary, MSAlign shared disentangle the candidates which facilitates retrieval ($E[d(m',m'')]=0.67$). \textbf{b)} Performances of in-batch and candidate InfoNCE when scaling the number of negatives i.e. the batchsize $B$ for in-batch and the hyperparameter $K$ for candidate InfoNCE. In the second case batchsize is fixed to $B=128$. \textbf{c)} Pairwise Jaccard index between the sets of candidates incorrectly ranked above the target by each scorer.}
    \label{fig:losses_figure}
    \vspace{-0.3cm}
\end{figure}

Results on MassSpecGym are reported in Table~\ref{tab:representations}. We observe that foundation models consistently outperform classical vectorization techniques: replacing spectral histograms with DreaMS improves performance by $+4.3\%$, replacing Morgan fingerprints with MolDeBERTa by $+10.2\%$, and combining both foundation models yields $+13.2\%$. In contrast, the choice among molecular foundation models has a comparatively limited impact: the different models perform similarly overall, with MolDeBERTa consistently achieving the best or second-best results in R@k scores.
\paragraph{Training loss.} MSAlign is trained with the candidate-based InfoNCE loss \eqref{eq:infoce_candidates}, we compare this approach against batch InfoNCE \eqref{eq:infoce_batch} and regression losses, $\mathcal{L}_\mathrm{reg}(\theta) = - \frac{1}{N} \sum_{i=1}^N \rho_\theta(s_i, m_i)$.
For regression, we consider both spectrum prediction, $\rho_\theta(s,m)=\operatorname{sim}(E_{ms}(s),h_\theta(m))$, and molecule prediction, $\rho_\theta(s,m)=\operatorname{sim}(g_\theta(s),E_{mol}(m))$. As shown in Table~\ref{tab:training_loss}, candidate-based InfoNCE substantially outperforms all alternatives. We attribute this gap to three observations:
\begin{itemize}[leftmargin=10pt, itemsep=0.2em, topsep=0pt, parsep=0pt]
    \item \textbf{Regression} (mol $\to$ ms): under-performs because this mapping is inherently one-to-many: in MassSpecGym, each molecule is associated with $19$ spectra on average. 
    \item \textbf{Regression} (ms $\to$ mol): is also challenging as candidates are tightly clustered in molecular space (Figure~\ref{fig:losses_figure}a), such that even small regression errors can substantially degrade retrieval accuracy.
    \item  \textbf{In-batch InfoNCE} trades informative candidate negatives for cheaper training. Figure~\ref{fig:losses_figure}b shows that this trade-off is unfavorable: candidate-based InfoNCE consistently outperforms its in-batch counterpart across batch sizes $B$ and numbers of negatives $K$.
\end{itemize}

\begin{table*}[t]
\vspace{-0.1cm}
\centering
\begin{minipage}[t]{0.50\textwidth}
\centering
\footnotesize
\caption{Training loss comparison on MassSpecGym Formula split (Average over 3 seeds).}
\vspace{-0.1cm}
\label{tab:training_loss}
\resizebox{!}{1.1cm}{%
\begin{tabular}{lccc}
\toprule
\multirow{2}{*}{\textbf{Training Loss}} & \multicolumn{3}{c}{\textbf{MassSpecGym}} \\
\cmidrule(lr){2-4}
& \textbf{R@1} & \textbf{R@5} & \textbf{R@20} \\
\midrule
Regression (mol $\rightarrow$ ms) & 22.8 {\scriptsize $\pm$ 2.1} & 42.0 {\scriptsize $\pm$ 1.3} & 60.3 {\scriptsize $\pm$ 0.6} \\
Regression (ms $\rightarrow$ mol) & 16.7 {\scriptsize $\pm$ 1.9} & 32.3 {\scriptsize $\pm$ 2.4} & 49.6 {\scriptsize $\pm$ 2.4} \\
Contrastive (in-batch) & 19.5 {\scriptsize $\pm$ 2.1} & 42.6 {\scriptsize $\pm$ 1.8} & 67.7 {\scriptsize $\pm$ 2.8} \\
\rowcolor{oursrow}
Contrastive (candidate-based) & \textbf{47.5} {\scriptsize $\pm$ 3.4} & \textbf{71.3} {\scriptsize $\pm$ 2.6} & \textbf{87.0} {\scriptsize $\pm$ 1.1} \\
\bottomrule
\end{tabular}
}
\end{minipage}
\hfill
\begin{minipage}[t]{0.45\textwidth}
\centering
\footnotesize
\vspace{-0.1cm}
\setlength{\tabcolsep}{4pt} 
\caption{Scorer ablation on MassSpecGym Formula split (Average over 3 seeds).}
\vspace{-0.1cm}
\label{tab:scorers}
\resizebox{!}{1.15cm}{%
\begin{tabular}{lccc}
\toprule
\multirow{2}{*}{\textbf{Model}} 
& \multicolumn{3}{c}{\textbf{MassSpecGym}} \\
\cmidrule(lr){2-4}
& \textbf{R@1} & \textbf{R@5} & \textbf{R@20} \\
\midrule

MSAlign 
& 47.5 
& 71.3 
& 87.0 \\

+ 3 Modalities 
& 54.4$^{\textcolor{green!60!black}{+6.9}}$
& 75.8$^{\textcolor{green!60!black}{+4.5}}$
& 89.9$^{\textcolor{green!60!black}{+2.9}}$ \\

+ Mass $\sigma \in \{0.01,0.1,1\}$
& 51.3$^{\textcolor{green!60!black}{+3.8}}$
& 76.7$^{\textcolor{green!60!black}{+5.4}}$
& 91.0$^{\textcolor{green!60!black}{+4.0}}$ \\

+ Both (\msalignplus)
& \textbf{58.2}$^{\textcolor{green!60!black}{+10.7}}$
& \textbf{80.7}$^{\textcolor{green!60!black}{+9.4}}$
& \textbf{92.8}$^{\textcolor{green!60!black}{+5.8}}$ \\

\bottomrule
\end{tabular}
}
\end{minipage}
\vspace{-0.3cm}
\end{table*}

\paragraph{Late score fusion.} In Section~\ref{sec:method}, we introduced
late score fusion as a simple yet effective approach to model aggregation and
defined \msalignplus \ as the fusion of $7$ complementary scorers: four
alignment models (including MSAlign) and several one-dimensional heuristics
based on the estimated mass of the target molecule. As shown in
Table~\ref{tab:main-results}, score fusion consistently improves R@1 by
approximately $10$\% over MSAlign, making \msalignplus \ the best-performing
method by a substantial margin. Importantly, this gain is not only due to
combining several equally strong models: Table~\ref{tab:scorers} shows that
MSAlign is itself the strongest individual scorer while the mass scorer are weak
on their own.

The ablation in Table~\ref{tab:scorers} further reveals a striking near-additivity of the gains: the alternative alignment models improve R@1 by $+6.9$\%, the mass-based scorers by $+3.8$\%, and combining both yields $+10.7$\%. To understand this complementarity, we analyze whether the different scorers make the same errors. For a scorer $\rho$, query spectrum $s\in\mathcal{S}$, and target molecule $m\in\mathcal{M}$, define the set of candidates incorrectly ranked above the target as $
C_\rho(s,m)
=
\left\{
m'\in C(s)
\;\middle|\;
\rho(s,m')>\rho(s,m)
\right\}$. For each pair of scorers $(\rho,\rho')$, we then compute the average Jaccard overlap between these sets,
$$
J(\rho,\rho')
=
\frac{1}{N}\sum_{i=1}^N
\frac{
\left|C_\rho(s_i,m_i)\cap C_{\rho'}(s_i,m_i)\right|
}{
\left|C_\rho(s_i,m_i)\cup C_{\rho'}(s_i,m_i)\right|
},
$$
where the terms with zero denominator are discarded from the average.

Results are reported in Figure~\ref{fig:losses_figure}c. Overall, the pairwise overlaps are small, particularly between alignment- and mass-based scorers. For instance, the overlap between MSAlign (DreaMS--MolDeBERTa) and the mass scorer with $\sigma=1$ is smaller than $5$\% . This confirms that the different scorers make largely complementary errors, providing a natural explanation for the effectiveness and near-additivity of late score fusion. Note that the Jaccard index can be somewhat misleading and we report a more detailed metric in Figure~\ref{fig:JaccardImproved}.

\begin{figure}[t!]
    \centering
    \includegraphics[width=0.48\linewidth]{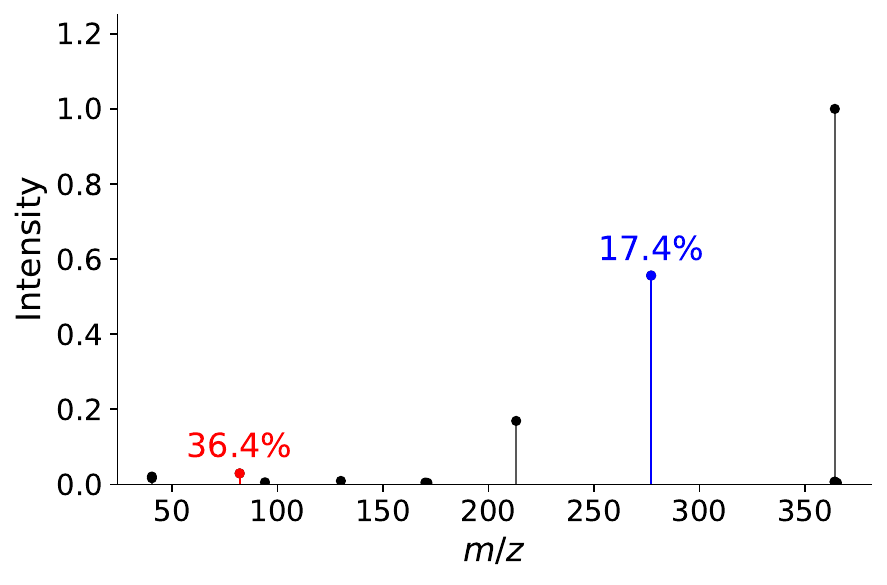}
    \includegraphics[width=0.48\linewidth]{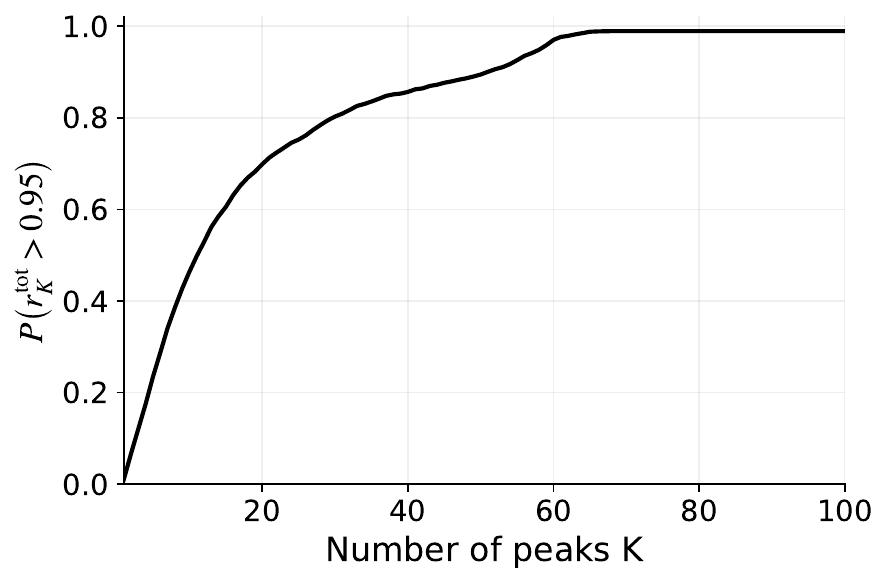}
    \caption{Using LIME to quantify peak importance. (Left) We illustrate the output of LIME by highlighting the two most important peaks of a spectra from MassSpecGym. (Right) We report the fraction samples (MassSpecGym test set) such that $K$ peaks are sufficient to explain MSAlign output.}
    \label{fig:lime}
    \vspace{-0.2cm}
\end{figure}

\subsection{Interpretability of peak importance with LIME}

MSAlign is a deep learning model with no built-in interpretability features which can be a limitation for practitioners. In this section, we demonstrate that this limitation can be mitigated by adapting \textbf{LIME}, a celebrated general purpose interpretability technique \cite{ribeiro2016should}, to our setting.

\paragraph{Adapating LIME to molecular retrieval.} LIME estimates individual feature contribution by fitting a linear regression under a user-defined distribution of feature perturbations. This approach can be smoothly adapted to molecular retrieval. Formally,
consider a sample $(s,m) \in \mathcal{S} \times \mathcal{M}$ where the spectra is composed of $n$ peaks: $s=(p_1,\ldots,p_n)$. We propose to explain the contribution of each peak to the scalar prediction margin
\begin{equation}
f(s)=\rho(s,m)-\max_{c\in\mathcal{C}\setminus\{m\}}\rho(s,c).
\end{equation}
To this end, we perturb the input spectra $s$ by randomly dropping some peaks: for any $z\in\{0,1\}^n$, we denote $s_z$ the spectra such that $z_i=1$ if and only if peak $p_i$ is dropped.
Then, following the general LIME strategy, we fit a linear regression to approximate $f(s)$ locally and the linear coefficient define the importance $r_i$ of peak $i$:
$$f(s_z)\approx\beta_0+\sum_{i=1}^n\beta_i z_i, \quad  r_i = \frac{|\beta_i|}{\sum_{j=1}^n |\beta_j|}.$$
In practice, $z$ is sampled from a Bernoulli distribution with parameter $q=0.1$ for each peak and the linear regression is fit using $n=1024$ samples. %This approach is illustrated in Figure~\ref{fig:lime} (Left) on a random spectra of MassSpecGym test set. We note that the most important peak (in red) is not the one with the largest intensity which is to be expected. 

\paragraph{Number of peaks contributing to MSAlign output.} To illustrate the interest of this method, we propose to quantify how many peaks are needed to support MSAlign prediction. To this end, we define $r_K^\mathrm{tot} = \sum_{i=1}^K r_i$, the total importance of the $K$ most important peaks (assuming $r_1>...>r_n$). For different values of $K$, we compute the fraction on spectra in MassSpecGym test set such that $r_K^\mathrm{tot} > 0.95$\% i.e. the fraction of samples such that $K$ peaks are sufficent to explain most of MSAlign prediction. The results are reported in Figure~\ref{fig:lime} (Right). Overall, we observe that MSAlign decision are based on at most $K=60$ peaks and require less that $K\leq 10$ for about half of the samples.

\section{Conclusion}
We addressed molecule retrieval from mass spectra and unify recent approaches within a single scoring framework. Inspired by vision--language alignment, we introduced MSAlign, a model that aligns pretrained spectrum and molecule foundation models within a shared space using candidate-based InfoNCE loss. Additionally, we presented a simple late score fusion strategy that significantly boosts the performances of the model. Finally, we contributed to the interpretability of the existing benchmarks by defining a practical metric to navigate the tradeoff between data leakage and distribution shift. A natural direction for future work is to attach a (possibly pre-trained) decoder to MSAlign shared space to tackle the challenging task of \textit{de novo} molecular identification. Pseudo-labeling with a forward model also appears as a key addition which should be explored in the future.

\begin{ack}
The study was funded by French National Research Agency (ANR) through the following projects: PEPR IA FOUNDRY (ANR-23-PEIA-0003), e-Lucid (ANR-25-TSIA-0002-01), MetaboHUB (ANR-11-INBS-0010) and through the
France 2030 program (ANR-23-IACL-0005) under the MacLeOD project (ANR-25-PEIA-0005).

It also received funding from the European Union’s Horizon Europe research and innovation programme under grant agreement 101120237 (ELIAS). 

The first  author received a PhD scholarship from Institut Polytechnique de Paris.
We thanks NeurIPS reviewers and AC for the fruitful rebuttal. Anonymous reviewer vvVF questions on mass filtering motivated the addition of mass based scorers and score fusion in general.
\end{ack}

\bibliographystyle{apalike}
{\small
\bibliography{references}}

%%%%%%%%%%%%%%%%%%%%%%%%%%%%%%%%%%%%%%%%%%%%%%%%%%%%%%%%%%%%

\appendix

\section{Comparison of in-batch and candidate-based InfoNCE \label{appendix:infonce}}

\subsection{In-batch InfoNCE is decorrelated from accuracy}

In-batch InfoNCE \eqref{eq:infoce_batch} leverages the other samples in the batch as negatives. We argue that, in a molecular setting, this strategy yields weak negatives, i.e. negatives that are easy to distinguish from the target. Consequently, a weak scorer can achieve a low in-batch infoNCE while performing poorly on the retrieval accuracy. 

Formally, for any score $\rho:\mathcal{S} \times \mathcal{M} \mapsto \mathbb{R}_+$ and batch of pairs $\mathcal{B} = (s_i,m_i)_{i=1}^B$, we define the \textit{in-batch} InfoNCE as,
\begin{equation}
\mathcal{L}_\text{batch}(\rho) \equiv - \frac{1}{B} \sum_{i=1}^B \log \frac{\exp(\rho(s_i, m_i))}{\sum_{j=1}^B\exp(\rho(s_i, m_j))}.
\label{eq:infoce_batch_generic}
\end{equation}
and the top-1 accuracy as,
\begin{equation}
    \mathrm{R@1}(\rho)
    \equiv\frac{1}{B}\sum_{i=1}^{B}\mathbf 1\!\left\{
    \argmax_{m'\in C_K(s_i)}\rho(s_i,m')=m_i\right\},
\label{eq:infonce_accuracy}
\end{equation}
and we claim that \eqref{eq:infoce_batch_generic} and \eqref{eq:infonce_accuracy} can become surprisingly decorrelated.

This can be illustrated by introducing a handcrafted score, defined solely from the masses of the target and candidate:
\begin{equation}
    \rho_{\lambda,\epsilon}(s,m) \equiv \lambda \ \mathbf{1}[ \mathrm{abs}(|m'| - |m_s|) < \epsilon ]
\end{equation}
where $|m_s|$ is the mass of target, which can be computed exactly if we assume access to the target's formula as metadata of the spectra $s$ which is the case for FLARE \citep{chen2026flare} for instance.

In practice, all candidates sets feature at least one candidate with the exact same mass, and thus the accuracy of this score is always zero,
\begin{equation}
        \mathrm{R@1}(\rho_{\lambda,\epsilon})= 0.
\end{equation}
Despite this catastrophic performance (below a random guess), we can show both empirically and theoretically that the \textit{in-batch} InfoNCE of the same scorer can go to zero.

\begin{figure}[h!]
    \centering
    \includegraphics[width=0.45\linewidth]{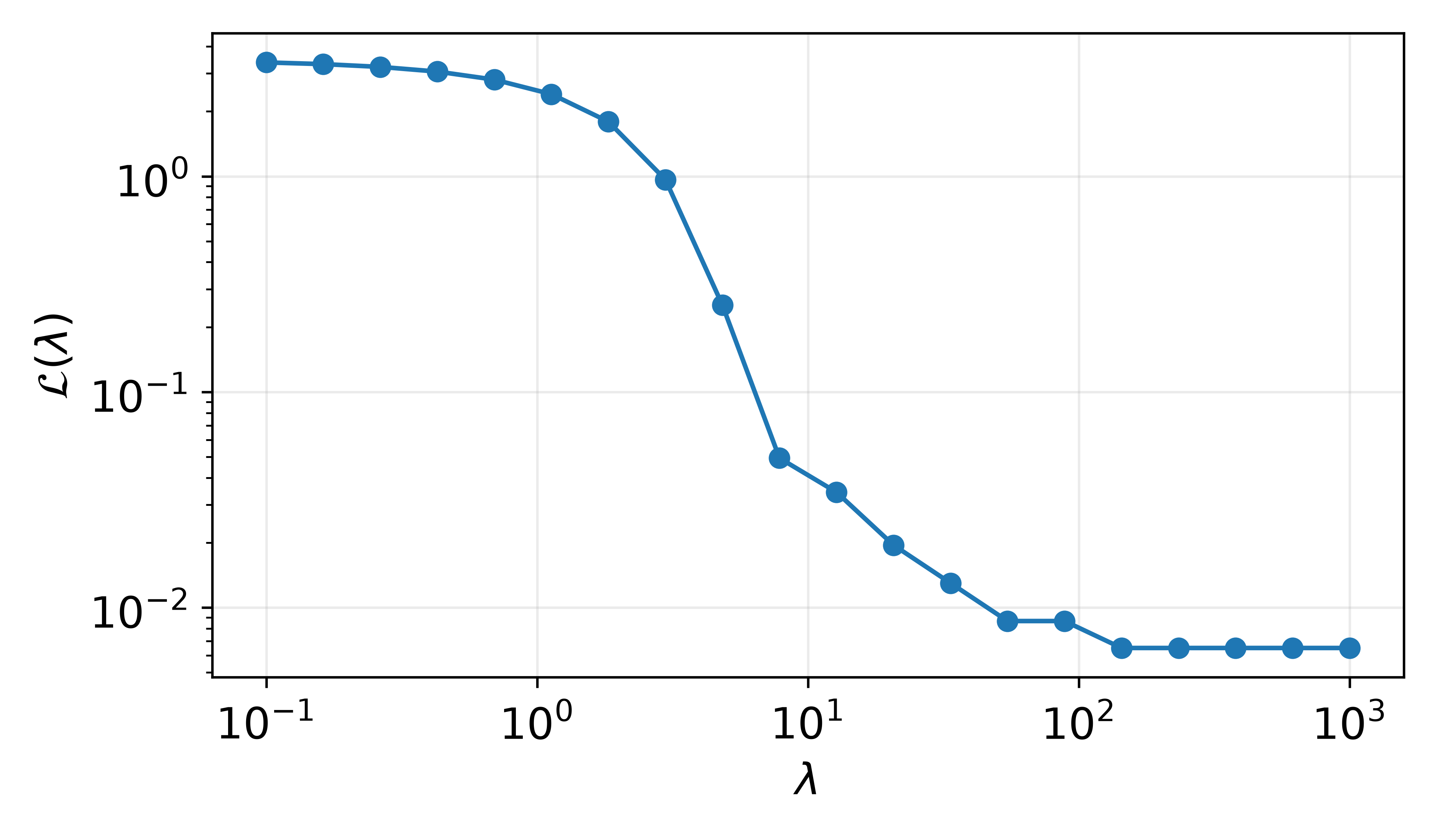}
    \hfill
    \includegraphics[width=0.45\linewidth]{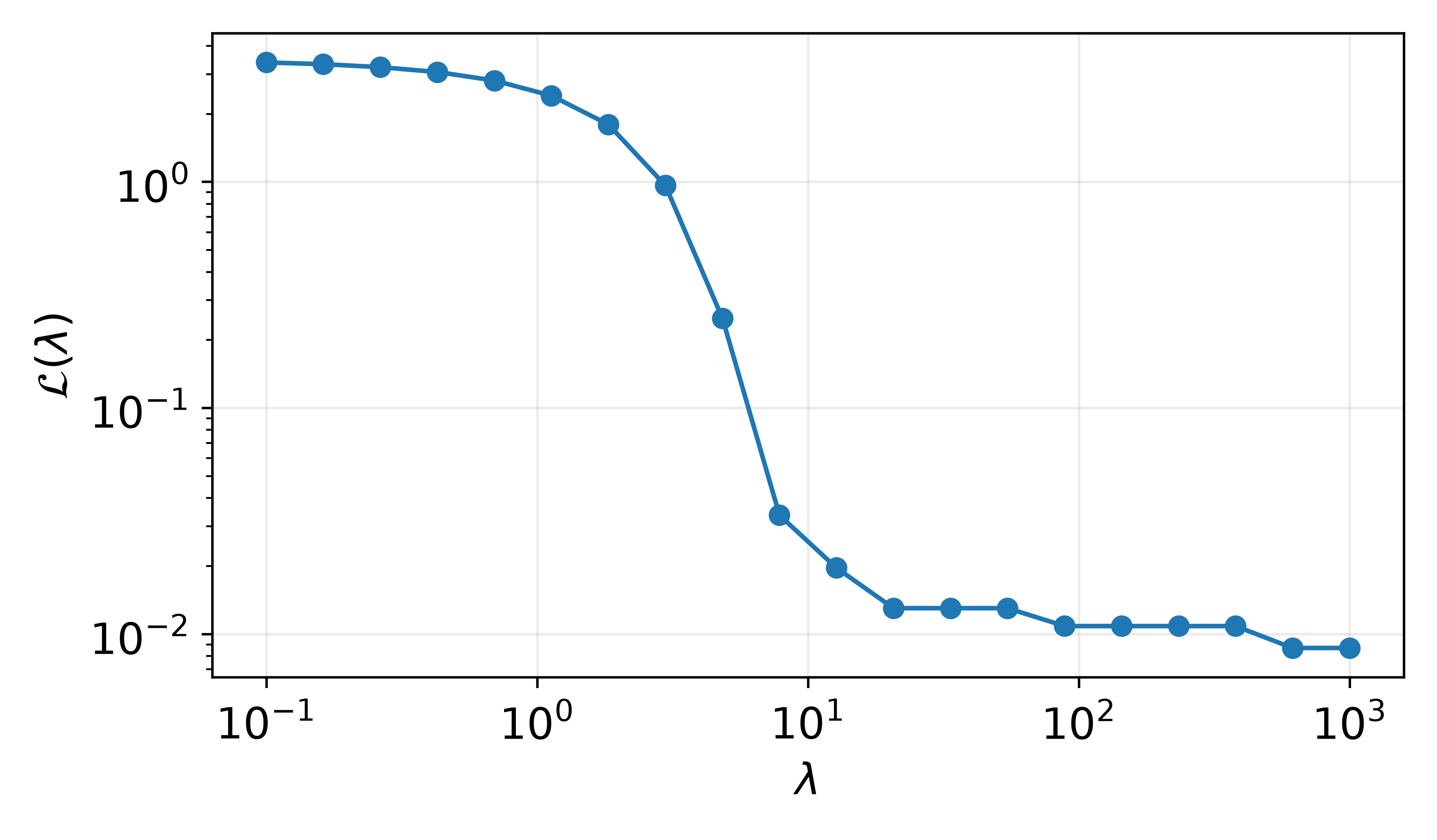}
    \caption{Expected in-batch InfoNCE loss \eqref{eq:inbatch_lambda} for MassSpecGym (Left) and SpectraVerse (Right). In both cases, the trivial scorer can achieve a loss below $10^{-2}$ while achieving 0$\%$ top-1 accuracy.}
    \label{fig:inbatch_loss}
\end{figure}

\paragraph{Empirical result.} Empirically, we set $\epsilon(\lambda) = \frac{1}{\lambda}$, and report the expected loss 
\begin{equation}
\label{eq:inbatch_lambda}
\mathcal{L}(\lambda) = E_\mathcal{B}[\mathcal{L}_\text{batch}(\rho_{\lambda,\epsilon(\lambda)})],
\end{equation}
on MassSpecGym and Spectraverse dataset at batchsize $B=32$ (Figure~\ref{fig:inbatch_loss}). In both case, we see that simply increasing $\lambda$ decrease the in-batch InfoNCE loss by several order of magnitude, while the accuracy remains null. 

\paragraph{Theoretical observation.} More generally, we can demonstrate that the trivial score $\rho_{\lambda,\epsilon}$ can always achieve low loss values and that this issue worsen when dataset size increase. 

\newtheorem{proposition}{Proposition}
\begin{proposition}[Upper bound of the in-batch loss]
Introduce the expected loss,

\begin{equation}
\label{eq:inbatch_lambdaeps}
\mathcal{L}(\lambda, \epsilon) = E_\mathcal{B}[\mathcal{L}_\text{batch}(\rho_{\lambda,\epsilon})],
\end{equation}

For any dataset we have,

\begin{equation}
\mathcal{L}(\lambda, \epsilon) \leq \log\left(1 + (B-1)(e^{-\lambda} (1-p_\epsilon) + p_\epsilon) \right)
\end{equation}

where $p_\epsilon$ is the probability for two random samples to have the same mass up to $\epsilon$. In particular, if the molecules in the dataset can take $N_m$ different and equiprobable masses, we have

\begin{equation}
\mathcal{L}(+\infty, 0) \leq \log\left(1 + \frac{B-1}{N_m-1} \right).
\end{equation}

Thus, by construction, the loss goes to zero when $N_m$ increases.
\end{proposition}
\begin{proof}
For a given batch we have,
\begin{equation*}
    \mathcal{L}_\text{batch}(\rho_{\lambda,\epsilon}) = \frac{1}{B} \sum_{i=1}^N \log\left(1 + \sum_{j \neq i} \exp(\rho_{\lambda,\epsilon}(s_i,m_j) - \rho_{\lambda,\epsilon}(s_i,m_i)) \right)
\end{equation*}
Where $\rho_{\lambda,\epsilon}(s_i,m_j) = \lambda \delta_{i,j}$ for 
\begin{equation*}
    \delta_{i,j} = \mathbf{1}[ \mathrm{abs}(|m_i| - |m_j|) < \epsilon ].
\end{equation*}
This rewrites as,
\begin{align*}
    \mathcal{L}_\text{batch}(\rho_{\lambda,\epsilon}) = \frac{1}{B} \sum_{i=1}^N \log\left(1 + \sum_{j \neq i} \exp(-\lambda)\exp(\lambda \delta_{i,j}) \right)
\end{align*}
Applying Jensen inequality we have,
\begin{align*}
    E_\mathcal{B}[\mathcal{L}_\text{batch}(\rho_{\lambda,\epsilon})] &\leq \frac{1}{B} \sum_{i=1}^N \log\left(1 + \sum_{j \neq i} \exp(-\lambda) E_\mathcal{B}[\exp(\lambda \delta_{i,j})] \right) \\
    &\leq \frac{1}{B} \sum_{i=1}^N \log\left(1 + \sum_{j \neq i} (B-1) (p_\epsilon + e^{-\lambda} (1-p_\epsilon) ] \right)\\
    &\leq  \log\left(1 + \sum_{j \neq i} (B-1) (p_\epsilon + e^{-\lambda} (1-p_\epsilon) ] \right).
\end{align*}
In particular $\mathcal{L}(+\infty, 0) = \log(1 + (B-1) p_\epsilon)$. Finally, in the hypothesis of a dataset with $N_m$ equiprobable masses, we have $p_\epsilon = \frac{1}{N_m - 1}$ which concludes the proof.
\end{proof}

Thus, increasing the number of unique molecular masses in the dataset is sufficient to guarantee that the in-batch InfoNCE loss of the trivial scorer goes to zero, even though the accuracy is still null.

We envision three options to prevent this behavior:
\begin{itemize}
    \item Option 1: increase the batchsize $B$. This is the simplest approach to make in-batch InfoNCE more informative, and in was successfully applied in language--vision alignment \citep{zhang2025sail,roschmann2026sotalign}. We tested this approach (with limtied success) in the experimental section, see Figure~\ref{fig:losses_figure}\textbf{b}.
    \item Option 2: hard negative mining \cite{robinson2020contrastive}. This is the approach we implement, using the candidate set as a cheap source of hard negatives.
    \item Option 3: batch sampling can be modified such that molecules that are similar are more likely to be sampled in the same batch. This approach is left for future works.
\end{itemize}

\subsection{Candidate-based InfoNCE is a valid surrogate loss for accuracy}

\label{appendix:msalign_top1_proof}
We first establish a scalar softmax bound, then apply it to the temperature-scaled candidate scores.

\newtheorem{lemma}{Lemma}
\begin{lemma}
\label{lem:softmax_top1}
For $x^+,x^-_1,\ldots,x^-_J\in\mathbb R$, with $J\geq1$, define
\begin{align*}
    \ell\big(x^+,\{x^-_k\}_{k=1}^{J}\big)
    &=-\log\frac{\exp(x^+)}{\exp(x^+)+\sum_{k=1}^{J}\exp(x^-_k)},\\
    \operatorname{err}\big(x^+,\{x^-_k\}_{k=1}^{J}\big)
    &=\mathbf 1\!\left\{\max_{1\leq k\leq J}x^-_k>x^+\right\}.
\end{align*}
Then
\begin{equation}
    \operatorname{err}\big(x^+,\{x^-_k\}_{k=1}^{J}\big)
    \leq\frac{\ell\big(x^+,\{x^-_k\}_{k=1}^{J}\big)}{\log 2}.
\label{eq:scalar_softmax_bound}
\end{equation}
\end{lemma}

\begin{proof}
Let $\Delta=\max_k x^-_k-x^+$. If $\Delta\leq0$, then $\operatorname{err}=0$ and the inequality follows from $\ell\geq0$. Otherwise, $\operatorname{err}=1$ and
\[
    \ell=\log\!\left(1+\sum_{k=1}^{J}\exp(x^-_k-x^+)\right)
    \geq\log(1+\exp(\Delta))>\log 2.
\]
This proves the strict-error bound. To obtain its non-strict version, observe that $\Delta=0$ also gives $\ell\geq\log 2$. Therefore,
\begin{equation}
    \mathbf 1\!\left\{\max_k x^-_k\geq x^+\right\}
    \leq\frac{\ell\big(x^+,\{x^-_k\}_{k=1}^{J}\big)}{\log 2}.
\label{eq:scalar_softmax_nonstrict}
\end{equation}
\end{proof}

\begin{proposition}
\label{prop:msalign_top1}
Define the empirical top-1 retrieval accuracy as
\begin{equation*}
    \mathrm{R@1}(\theta)
    =\frac{1}{B}\sum_{i=1}^{B}\mathbf 1\!\left\{
    \argmax_{m'\in C_K(s_i)}\rho_\theta(s_i,m')=m_i\right\}.
\end{equation*}
The candidate-based InfoNCE loss satisfies
\begin{empheq}[box=\fbox]{equation*}
    1-\mathrm{R@1}(\theta)
    \leq\frac{\mathcal L_{\mathrm{cand}}(\theta)}{\log 2}.
\end{empheq}
\end{proposition}

\begin{proof}
Apply Lemma~\ref{lem:softmax_top1} with $x^+=\rho_\theta(s_i,m_i)/\epsilon$ and negative scores $x^-_k=\rho_\theta(s_i,m')/\epsilon$ for $m'\in C_K(s_i)\setminus\{m_i\}$. Since $\epsilon>0$, the ranking is unchanged and the scalar loss equals $\ell_i(\theta)$. Let $\widehat m_i=\argmax_{m'\in C_K(s_i)}\rho_\theta(s_i,m')$ denote the predicted molecule. An incorrect prediction implies that a negative scores at least as high as the target. Hence, by~\eqref{eq:scalar_softmax_nonstrict},
\[
    \mathbf 1\{\widehat m_i\neq m_i\}
    \leq\mathbf 1\!\left\{\max_{m'\in C_K(s_i)\setminus\{m_i\}} \rho_\theta(s_i,m')\geq\rho_\theta(s_i,m_i)\right\}
    \leq\frac{\ell_i(\theta)}{\log 2}.
\]
Averaging over queries gives
\[
    1-\mathrm{R@1}(\theta)
    \leq\frac{1}{B\log 2}\sum_{i=1}^{B}\ell_i(\theta)
    =\frac{\mathcal L_{\mathrm{cand}}(\theta)}{\log 2}.
\]
\end{proof}
\section{Implementation Details \label{appendix:details}}

\subsection{Pre-processing and training.}

\begin{table}[!h]
\centering
\caption{Dataset statistics.}
\vspace{2mm}
\label{tab:dataset_stats}
\begin{tabular}{lcc}
\toprule
& \textbf{MassSpecGym} & \textbf{Spectraverse} \\
\midrule
\textbf{\# Mol/MS pairs}  & 231,104 & 488,797 \\
\textbf{\# Unique molecules}  & 28,936 & 44,307 \\
\textbf{Non [M+H]+} & 15\% & 52\% \\
\bottomrule
\end{tabular}
\end{table}

\paragraph{Molecule pre-processing.} All SMILES strings are canonicalized and sanitized using RDKit~\citep{landrum2013rdkit}. Chemical formulas and weights are computed by explicitly accounting for implicit hydrogen atoms. The molecular masses are computed with RDKit ExactMolWt function which uses the monoisotopic atomic mass (mass of the most abundant isotope). All baselines are trained using the hyperparameters reported in their respective original works. 

\paragraph{Model training.} MSAlign is trained on a single NVIDIA V100 GPU, using the AdamW optimizer with linear warmup followed by cosine decay. Following common practice, we augment the contrastive losses with a temperature parameter $\epsilon$, for instance $\mathcal{L}_\text{batch}$ writes as:
\begin{equation}
\mathcal{L}_\text{batch}(\theta) = - \sum_{i=1}^B \log \frac{\exp(\rho_\theta(s_i, m_i)/\epsilon)}{\sum_{j=1}^B\exp(\rho_\theta(s_i, m_j)/\epsilon)}.
\end{equation}
and $\epsilon$ is a trainable parameters parametrized in log scale and initialized to $0.07$.

\begin{table}[H]
\centering
\caption{Main MSAlign hyperparameters.}
\label{tab:msalign-hyperparameters}
\small
\resizebox{\textwidth}{!}{%
\begin{tabular}{lccc}
\toprule
\textbf{Hyperparameter}
& \textbf{MassSpecGym MCES}
& \textbf{MassSpecGym Formula}
& \textbf{SpectraVerse Formula} \\
\midrule
MS representation
& DreaMS
& DreaMS
& DreaMS \\
Molecular representation
& MolDeBERTa
& MolDeBERTa
& MolDeBERTa \\
Shared dimension $d_{\mathrm{model}}$
& 512 & 512 & 512 \\
Metadata dimension $d_{\mathrm{meta}}$
& 64 & 64 & 64 \\
% Metadata conditioning
% & Concatenation & Concatenation & Concatenation \\
% Collision-energy encoder
% & Fourier (32 frequencies) & Fourier (32 frequencies) & Fourier (32 frequencies) \\
% Adduct encoding
% & Learned, sign-aware & Learned, sign-aware & Learned, sign-aware \\
\midrule
% MS pre-fusion MLP hidden layers
% & 0 & 0 & 0 \\
MS MLP hidden layers
& 1 & 2 & 2 \\
Molecule MLP hidden layers
& 3 & 3 & 3 \\
MLP hidden dimension
& 2048 & 2048 & 2048 \\
Activation / normalization
& GELU / LayerNorm & GELU / LayerNorm & GELU / LayerNorm \\
MLP dropout
& 0.3 & 0.3 & 0.2 \\
% Residual shortcut
% & Yes & Yes & Yes \\
\midrule
Loss
% & Candidate cross-entropy
% & Candidate cross-entropy
% & Candidate cross-entropy \\
Score temperature
& Learned ($0.07$ init.)
& Learned ($0.07$ init.)
& Learned ($0.07$ init.) \\
Candidates per query
& 128 & 128 & 128 \\
Batch size
& 128 & 128 & 128 \\
Learning rate
& $10^{-4}$ & $10^{-4}$ & $10^{-4}$ \\
Weight decay
& $10^{-6}$ & $10^{-6}$ & $10^{-6}$ \\
Warmup steps
& 2,000 & 2,000 & 2,000 \\
Maximum training steps
& 20,000 & 30,000 & 50,000 \\
\bottomrule
\end{tabular}}
\end{table}

\subsection{Late score fusion.}
\label{appendix:fusion_algorithm}
\paragraph{Smooth surrogate.}
The validation top-1 accuracy is defined as 
\[
    \mathrm{R@1}_{\mathrm{val}}(\boldsymbol\alpha)
    =\frac{1}{|\mathcal D_{\text{val}}|}
    \sum_{(s_i,m_i)\in\mathcal D_{\text{val}}}
    \mathbf 1\!\left\{\argmax_{m'\in C(s_i)}
    \rho_{\boldsymbol\alpha}(s_i,m')=m_i\right\}.
\]
To approximate this objective, we maximize a concave log-softmax surrogate over the simplex for each fixed fusion temperature $T>0$:
\begin{equation}
    \boldsymbol\alpha_T
    \in\argmax_{\boldsymbol\alpha\in\Delta^{L-1}}
    \frac{1}{|\mathcal D_{\text{val}}|}
    \sum_{(s_i,m_i)\in\mathcal D_{\text{val}}}
    \log\frac{\exp(\rho_{\boldsymbol\alpha}(s_i,m_i)/T)}
    {\sum_{m'\in C(s_i)}\exp(\rho_{\boldsymbol\alpha}(s_i,m')/T)}.
\label{eq:msalign_listwise_calibration}
\end{equation}
This is a convex optimization problem. Here, $T$ is distinct from the learned training temperature $\epsilon$. The component scores remain fixed, and each full candidate set $C(s_i)$ includes the target $m_i$.

\paragraph{Temperature selection.}
The surrogate need not share the maximizer of R@1. We therefore fit weights for each temperature in a finite grid $\mathcal T\subset(0,\infty)$ and select them using the original validation metric:
\begin{equation}
    T^\star\in\argmax_{T\in\mathcal T}
    \mathrm{R@1}_{\mathrm{val}}(\boldsymbol\alpha_T),
    \qquad
    \widehat{\boldsymbol\alpha}=\boldsymbol\alpha_{T^\star}.
\label{eq:msalign_fusion_temperature}
\end{equation}
Both stages use validation data; the test split is reserved for evaluation. This procedure approximates the direct R@1 objective without guaranteeing its global optimum. Component scores are cached once and reused across temperatures, so fitting requires no further model evaluations. Algorithm~\ref{alg:fusion_calibration} summarizes the procedure. At inference, rank candidates by $\rho_{\widehat{\boldsymbol\alpha}}$. The fusion temperature is used only to fit the weights.

\newcounter{fusionalgorithm}
\begin{figure}[H]
\begin{minipage}{\linewidth}
\refstepcounter{fusionalgorithm}\label{alg:fusion_calibration}
\hrule\medskip
\textbf{Algorithm \thefusionalgorithm: Late fusion with temperature selection}
\medskip\hrule\medskip
\textbf{Input:} Fixed score functions $\rho_1,\ldots,\rho_L$; validation split
$\mathcal D_{\text{val}}$; full candidate sets $C(s_i)$ containing $m_i$;
fusion-temperature grid $\mathcal T\subset(0,\infty)$.\\
\textbf{Output:} Fusion weights $\widehat{\boldsymbol\alpha}\in\Delta^{L-1}$.
\begin{enumerate}[leftmargin=*,itemsep=0.4em]
    \item Cache $S_{i\ell}(m')\gets\rho_\ell(s_i,m')$ for every validation pair $(s_i,m_i)$, candidate $m'\in C(s_i)$, and component $\ell$.
    \item For each $T\in\mathcal T$:
    \begin{enumerate}[label=(\alph*),leftmargin=*,itemsep=0.2em]
        \item Solve~\eqref{eq:msalign_listwise_calibration} for $\boldsymbol\alpha_T$ using the cache and a convex solver over the simplex. Evaluate the objective with a numerically stable log-sum-exp.
        \item Compute $r_T\gets\mathrm{R@1}_{\mathrm{val}}(\boldsymbol\alpha_T)$ from the cached weighted scores, using the fixed tie-breaking rule.
    \end{enumerate}
    \item Select $T^\star\in\argmax_{T\in\mathcal T}r_T$ and return $\widehat{\boldsymbol\alpha}=\boldsymbol\alpha_{T^\star}$.
\end{enumerate}
\medskip\hrule
\end{minipage}
\end{figure}

\subsection{MSAlign architecture}

\paragraph{MLPs.} The main workhorse of the proposed alignement models are Multi Layers Perceptrons (MLPs). In this work, a MLP with \(L\) hidden layers applies an affine map, layer normalization, GELU, and dropout in each hidden block, followed by an affine output projection:
\[
h_0=x,\qquad
h_\ell=\operatorname{Dropout}\!\left(\operatorname{GELU}\!\left(\operatorname{LayerNorm}(W_\ell h_{\ell-1}+b_\ell)\right)\right),\qquad
\operatorname{MLP}_L(x)=W_{\mathrm{out}}h_L+b_{\mathrm{out}}.
\]

\paragraph{Adduct encoding.} Query spectra often contains meta-data such as the ionization mode which is a presented as a discrete adduct class $a \in \{1,...,N_a\}$. The adduct encoder concatenates a learned class embedding with a learned polarity embedding, allocating half of its 64-dimensional output to each:
\[
\phi_a(a)=\big[E_{\mathrm{class}}(a)\,;\,E_{\mathrm{sign}}(\sigma(a))\big]\in\mathbb{R}^{64}.
\]
Unknown or dropped adducts use a dedicated unknown class and unknown-sign embedding.

\paragraph{Collision energy.} Similarly, the collision energy uses a fixed 64-dimensional Fourier encoding, with 32 sine–cosine pairs applied to the scalar normalized by \(e_{\max}=100\):
\[
\phi_e(e)_{2j:2j+2}
=
\left[
\sin\left(\frac{e/e_{\max}}{b^{j/32}}\right),
\cos\left(\frac{e/e_{\max}}{b^{j/32}}\right)
\right],\qquad j=0,\ldots,31,\quad b=10^4.
\]
A missing energy value is represented by a separate learned vector.

\paragraph{Full architecture.}
The spectrum representation passes through a linear input projection, is concatenated with both metadata encodings, and passes through an output MLP; the molecular representation has its own MLP. Both towers produce 512-dimensional vectors:
\[
z_s=\operatorname{MLP}_{\mathrm{ms,out}}\!\left(
\big[\operatorname{Linear}_{\mathrm{ms,in}}(s)\,;\,\phi_e(e)\,;\,\phi_a(a)\big]\right),
\qquad
z_m=\operatorname{MLP}_{\mathrm{mol}}(m).
\]

\section{Additional Results \label{appendix:additionnal_results}}

\begin{figure}[t]
    \centering
    \includegraphics[width=\linewidth]{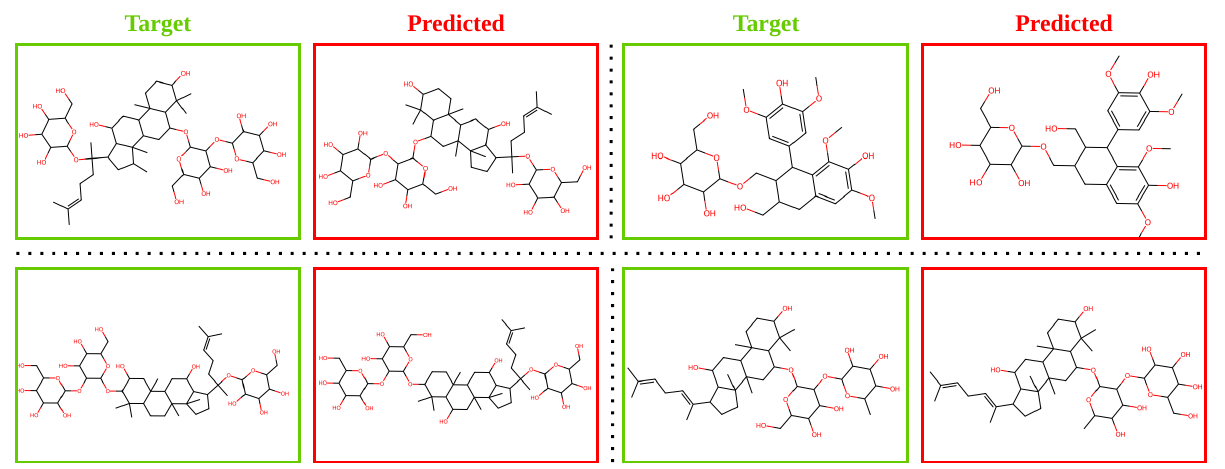}
    \caption{Four MSAlign failures cases, in each case we report both the prediction (in red) and the target (in green). The displayed failure cases are randomly sampled from the $5$\% hardest samples, in the sense that the Tanimoto distance between the target and closest candidate is the smallest.}
    \label{fig:failure_cases}
\end{figure}

\subsection{MSAlign failure cases}

We observe that MSAlign failures are often due to a particularly challenging set of candidates i.e. when at least one of the candidate is particularly similar to the target.

To quantify this phenomenon, we propose to define the difficulty of a sample $(s,m) \in \mathcal{S} \times \mathcal{M}$ with candidate set $C(s) \subset \mathcal{M}$ as:
\begin{equation}
    \mathrm{Difficulty}(s) =  \min_{c \in C(s)} d(m,c)
    \label{eq:difficutly}
\end{equation}
where $d$ is distance over the set of molecules such as the (negative) MCES or Tanimoto similarity. Next we propose to restrict the test dataset the top-$K$\% of most difficult samples and compute the accuracy on this subset. Results are reported in Table~\ref{tab:failurecases} and we observe a correlation between the difficulty of the sample as defined in \eqref{eq:difficutly} and MSAlign performances.

\begin{table}[h!]
\centering
\caption{MSAlign R@1 accuracy on increasingly difficult subsets of MassSpecGym Formula-split test set. The subsets are defined by selecting the top-$K$\% of most difficult samples as defined by \eqref{eq:difficutly}. We report the results both for MCES and Tanimoto metrics.}
\label{tab:failurecases}
\vspace{0.1cm}
\resizebox{\textwidth}{!}{%
\begin{tabular}{lccccccc}
\toprule
\textbf{K} & 100\% & 80\% & 60\% & 40\% & 20\% & 10\% & 5\% \\
\midrule
\textbf{MCES}
& 47.52 ± 3.39
& 46.25 ± 4.10
& 45.83 ± 5.68
& 43.45 ± 7.57
& 35.26 ± 5.67
& 22.34 ± 6.76
& 21.96 ± 14.88 \\
\textbf{Tanimoto}
& 47.52 ± 3.39
& 46.06 ± 4.18
& 42.93 ± 4.34
& 37.75 ± 8.30
& 29.31 ± 5.33
& 24.02 ± 9.90
& 25.17 ± 17.33 \\
\bottomrule
\end{tabular}}
\end{table}

\subsection{Methods training time comparison.}

\begin{table}[h!]
\centering
\caption{Training time in hours.min for the different methods on a single NVIDIA V100 GPU. }
\label{tab:train_time}
\vspace{0.1cm}
\resizebox{\textwidth}{!}{%
\begin{tabular}{lcccccccc}
\toprule
\textbf{Dataset}
& \textbf{DeepSet}
& \textbf{JESTR}
& \textbf{Emb-Cos}
& \textbf{MIST}
& \textbf{MVP}
& \textbf{FLARE}
& \textbf{MSAlign}
& \textbf{\msalignplus} \\
\midrule
MassSpecGym Formula
& 0.65 & 7.34 & 4.45 & 2.30 & 16.53 & 2.50 & 0.82 & 3.29 \\
MassSpecGym MCES
& 0.76 & 9.18 & 4.58 & 1.05 & 15.00 & 2.22 & 0.75 & 3.02 \\
SpectraVerse Formula
& 1.43 & 13.32 & 4.56 & 3.1. & 25.00 & 4.17 & 1.31 & 5.71 \\
\bottomrule
\end{tabular}}
\end{table}
We report the training time of the different methods in Table~\ref{tab:train_time}. Note that the training time reported for MSAlign$^*$ includes the training of the 4 independent alignment models (DreaMS/MolDeBERTa, DreaMS/Fingerprint...) and can be largely parallelized. The pre-computation of MolDeBERTa and DreaMS embedding takes about 5 hours but it needs to be run only once for the full project and is amortized across runs, seeds and models. Note that JESTR, FLARE and MVP are particularly slow despite using the cheaper candidate-based InfoNCE. This is due to the large number of epochs (1000 for JESTR) recommended in the original papers 
\cite{kalia2025jestr,chen2026flare,zhou2026learning}.

\subsection{Additionnal results on score fusion.}

\paragraph{Scorers ablation.} In Table~\ref{tab:scorers-long} we report the performances of each individual scorer used in \msalignplus \ and different combinations of scorers aggregations. In particular, we observe that MSAlign (DreaMS/MolDeBERTa model) is the best performing of all the individual scorers while on the contrary the 1-dimensional mass scorers achieve very poor performances on their own. 

\begin{table*}[h!]
\centering
\caption{Individual scorers and calibrated scorer combinations. B: binned spectra; D: DreaMS; FP: Morgan fingerprint; MD: MolDeBERTa. Mass columns give the Gaussian scorer scale in ppm. R@1 values are mean $\pm$ sample standard deviation over the three MassSpecGym formula test splits.}
\label{tab:scorers-long}
\resizebox{\textwidth}{!}{%
\begin{tabular}{lccccccc|c}
\toprule
\textbf{Configuration} & \textbf{B/FP} & \textbf{B/MD} & \textbf{D/FP} & \textbf{D/MD} & \textbf{Mass .01} & \textbf{Mass .1} & \textbf{Mass 1} & \textbf{R@1} \\
\midrule
\multirow{3}{*}{Alternative Modality Pair} & \textcolor{green!60!black}{\checkmark} & \textcolor{red!70!black}{\ensuremath{\times}} & \textcolor{red!70!black}{\ensuremath{\times}} & \textcolor{red!70!black}{\ensuremath{\times}} & \textcolor{red!70!black}{\ensuremath{\times}} & \textcolor{red!70!black}{\ensuremath{\times}} & \textcolor{red!70!black}{\ensuremath{\times}} & 34.3 $\pm$ 1.3 \\
 & \textcolor{red!70!black}{\ensuremath{\times}} & \textcolor{green!60!black}{\checkmark} & \textcolor{red!70!black}{\ensuremath{\times}} & \textcolor{red!70!black}{\ensuremath{\times}} & \textcolor{red!70!black}{\ensuremath{\times}} & \textcolor{red!70!black}{\ensuremath{\times}} & \textcolor{red!70!black}{\ensuremath{\times}} & 44.5 $\pm$ 5.1 \\
 & \textcolor{red!70!black}{\ensuremath{\times}} & \textcolor{red!70!black}{\ensuremath{\times}} & \textcolor{green!60!black}{\checkmark} & \textcolor{red!70!black}{\ensuremath{\times}} & \textcolor{red!70!black}{\ensuremath{\times}} & \textcolor{red!70!black}{\ensuremath{\times}} & \textcolor{red!70!black}{\ensuremath{\times}} & 38.6 $\pm$ 2.8 \\
\rowcolor{oursrow}
 MSAlign & \textcolor{red!70!black}{\ensuremath{\times}} & \textcolor{red!70!black}{\ensuremath{\times}} & \textcolor{red!70!black}{\ensuremath{\times}} & \textcolor{green!60!black}{\checkmark} & \textcolor{red!70!black}{\ensuremath{\times}} & \textcolor{red!70!black}{\ensuremath{\times}} & \textcolor{red!70!black}{\ensuremath{\times}} & 47.5 $\pm$ 3.4 \\
\midrule
MSAlign + Other modalities & \textcolor{green!60!black}{\checkmark} & \textcolor{green!60!black}{\checkmark} & \textcolor{green!60!black}{\checkmark} & \textcolor{green!60!black}{\checkmark} & \textcolor{red!70!black}{\ensuremath{\times}} & \textcolor{red!70!black}{\ensuremath{\times}} & \textcolor{red!70!black}{\ensuremath{\times}} & \underline{54.4} $\pm$ 3.0 \\
\midrule
\multirow{3}{*}{Gaussian Mass Scorer} & \textcolor{red!70!black}{\ensuremath{\times}} & \textcolor{red!70!black}{\ensuremath{\times}} & \textcolor{red!70!black}{\ensuremath{\times}} & \textcolor{red!70!black}{\ensuremath{\times}} & \textcolor{green!60!black}{\checkmark} & \textcolor{red!70!black}{\ensuremath{\times}} & \textcolor{red!70!black}{\ensuremath{\times}} & 0.6 $\pm$ 0.1 \\
 & \textcolor{red!70!black}{\ensuremath{\times}} & \textcolor{red!70!black}{\ensuremath{\times}} & \textcolor{red!70!black}{\ensuremath{\times}} & \textcolor{red!70!black}{\ensuremath{\times}} & \textcolor{red!70!black}{\ensuremath{\times}} & \textcolor{green!60!black}{\checkmark} & \textcolor{red!70!black}{\ensuremath{\times}} & 0.8 $\pm$ 0.2 \\
 & \textcolor{red!70!black}{\ensuremath{\times}} & \textcolor{red!70!black}{\ensuremath{\times}} & \textcolor{red!70!black}{\ensuremath{\times}} & \textcolor{red!70!black}{\ensuremath{\times}} & \textcolor{red!70!black}{\ensuremath{\times}} & \textcolor{red!70!black}{\ensuremath{\times}} & \textcolor{green!60!black}{\checkmark} & 0.7 $\pm$ 0.2 \\
\midrule
MSAlign + Mass & \textcolor{red!70!black}{\ensuremath{\times}} & \textcolor{red!70!black}{\ensuremath{\times}} & \textcolor{red!70!black}{\ensuremath{\times}} & \textcolor{green!60!black}{\checkmark} & \textcolor{green!60!black}{\checkmark} & \textcolor{green!60!black}{\checkmark} & \textcolor{green!60!black}{\checkmark} & 51.3 $\pm$ 3.4 \\
\midrule
\rowcolor{oursrow}
MSAlign$^{4M}$ & \textcolor{green!60!black}{\checkmark} & \textcolor{green!60!black}{\checkmark} & \textcolor{green!60!black}{\checkmark} & \textcolor{green!60!black}{\checkmark} & \textcolor{green!60!black}{\checkmark} & \textcolor{green!60!black}{\checkmark} & \textcolor{green!60!black}{\checkmark} & \textbf{58.2} $\pm$ 3.4 \\
\bottomrule
\end{tabular}%
}
\end{table*}

\paragraph{Scorer failure-case overlaps.}
In Figure~\ref{fig:losses_figure}c, we reported the pairwise Jaccard index between the failure sets of the individual scorers, defined for a scorer $\rho$, query spectrum $s$, and target molecule $m$ as
\begin{equation} C_\rho(s,m) = \left\{ m'\in C(s) \;\middle|\; \rho(s,m')>\rho(s,m) \right\}. \end{equation}
In other words, $C_\rho(s,m)$ contains all candidate molecules incorrectly ranked above the target by $\rho$.

\begin{figure}[h!]
    \centering
    \includegraphics[width=0.7\linewidth]{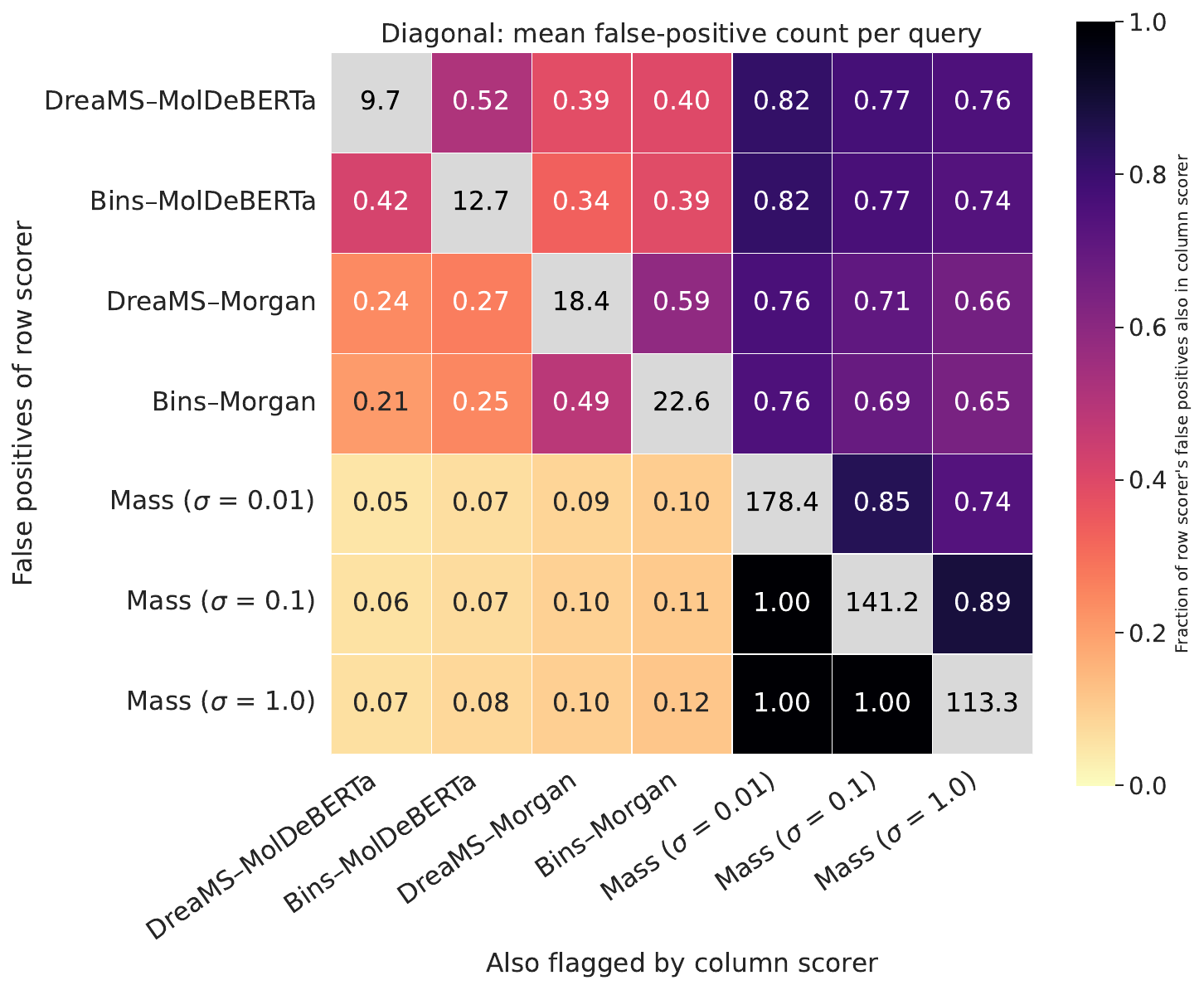}
    \caption{Average fraction of candidates incorrectly ranked by $\rho$ (rows) that also fool scorer $\rho'$ (column). The diagonal reports the average number of incorrectly ranked candidates for each individual scorer. MSAlign is the first scorer (DreaMS-MolDeBERTa) and \msalignplus\ is the aggregate of all scorers via late score fusion.}
    \label{fig:JaccardImproved}
\end{figure}

While the Jaccard index provides a symmetric measure of overlap, it can be difficult to interpret when the failure sets have substantially different sizes. A small Jaccard index may reflect genuinely complementary errors, but can also result from one failure set being much larger than the other. To obtain a more fine-grained picture, we therefore consider the following asymmetric overlap:
$$ J'(\rho,\rho') = \frac{1}{N}\sum_{i=1}^N \frac{ \left|C_\rho(s_i,m_i)\cap C_{\rho'}(s_i,m_i)\right| }{ \left|C_\rho(s_i,m_i)\right| }. $$
$J'(\rho,\rho')$ can be interpreted as the average fraction of failure cases of $\rho$ that also fool $\rho'$. Unlike the Jaccard index, it is asymmetric: $J'(\rho,\rho')$ and $J'(\rho',\rho)$ need not coincide. All pairwise values are reported in Table~\ref{fig:JaccardImproved}.

For completeness, we also report the average size of $C_\rho(s_i,m_i)$ for each scorer. Interestingly, these failure-set sizes exhibit a relatively heavy-tailed distribution: a scorer may correctly rank the target first for a large fraction of queries while ranking it below many candidates on a smaller number of difficult queries. For instance, for MSAlign (DreaMS--MolDeBERTa), we obtain
\begin{align} &\text{R@1} = \frac{1}{N}\sum_{i=1}^N \mathbf{1}[\left|C_\rho(s_i,m_i)\right|) = 0] = 47.5\% \\ &\text{Average} = \frac{1}{N}\sum_{i=1}^N \left|C_\rho(s_i,m_i)\right| = 9.7 \end{align}
Thus, despite ranking the correct molecule first for nearly half of the queries, MSAlign has an average of $9.7$ candidates ranked above the target, illustrating the strong heterogeneity of failure severity across queries which also contributes to the efficiency of model aggregation.

\section{Existing criterion for ms/ms dataset data splitting. \label{appendix:splits}}

Data splitting requires particular care in MS/MS benchmarks, as the same molecule may be associated with multiple spectra acquired under different experimental conditions. A naive random split can therefore place spectra of the same, or highly similar, molecules in both training and test sets, leading to data leakage and overly optimistic performance estimates. To mitigate this issue, several splitting strategies have been proposed, differing in the notion and degree of molecular similarity that they enforce across folds:

\begin{itemize}[leftmargin=15pt]
\setlength\itemsep{0.3em}
\setlength{\parskip}{0pt}
\setlength{\topsep}{-5pt}
\item \textbf{InChIKey split.} The InChIKey provides a unique identifier for a molecule's 2D structure. Splitting by InChIKey ensures that the same molecular structure does not appear in different folds~\cite{goldman2023mist}.

\item \textbf{Formula split.} Similarly, molecules are grouped according to their molecular formula, ensuring that molecules with the same formula do not appear in different folds. This criterion is strictly more restrictive than the InChIKey split: molecules with the same InChIKey necessarily have the same molecular formula, whereas distinct molecular structures may share the same formula.

\item \textbf{Murcko scaffold split.} Molecules are grouped according to their Murcko scaffold~\cite{landrum2013rdkit}, ensuring that molecules sharing the same underlying scaffold do not appear in different folds.

\item \textbf{Murcko histogram split.} DreaMS~\cite{dreams} introduces a stronger variant of the Murcko scaffold split, in which even a coarse histogram representation of the Murcko scaffold must differ across folds.

\item \textbf{MCES split.} MassSpecGym~\cite{massspecgym} uses the Maximum Common Edge Subgraph (MCES) distance to explicitly constrain structural similarity across folds, requiring an edit distance of at least $10$ between molecules assigned to different folds.
\end{itemize}

These strategies do not form a single hierarchy of increasingly restrictive splits. The InChIKey split is strictly less restrictive than all the alternatives above, while the Murcko histogram split is strictly more restrictive than the standard Murcko scaffold split. Beyond these relations, the criteria are generally incomparable: for instance, separating molecules by molecular formula, scaffold, or MCES distance captures different notions of structural similarity. Consequently, there is no obvious way to rank these strategies by their strength, making it difficult to make a principled choice between them.

%\newpage
%\include{Source/problem-background}

%%%%%%%%%%%%%%%%%%%%%%%%%%%%%%%%%%%%%%%%%%%%%%%%%%%%%%%%%%%%

\newpage
\section*{NeurIPS Paper Checklist}

\begin{enumerate}

\item {\bf Claims}
    \begin{itemize}
        \item[] Question: Do the main claims made in the abstract and introduction accurately reflect the paper's contributions and scope?
        \item[] Answer: \answerYes{} % Replace by \answerYes{}, \answerNo{}, or \answerNA{}.
        \item[] Justification: Yes all claims are justified and in particular we provide detailed ablations studies.
    \end{itemize}

\item {\bf Limitations}
    \item[] Question: Does the paper discuss the limitations of the work performed by the authors?
    \item[] Answer: \answerYes{} % Replace by \answerYes{}, \answerNo{}, or \answerNA{}.
    \item[] Justification: Yes we explcitly state that our works relies on the quality of the underlying foundation models and that the proposed approach do not apply to de novo structure elucidation.

\item {\bf Theory assumptions and proofs}
    \item[] Question: For each theoretical result, does the paper provide the full set of assumptions and a complete (and correct) proof?
    \item[] Answer: \answerYes{}{}.
    \item[] Yes, the proof are provided and correct.

    \item {\bf Experimental result reproducibility}
    \item[] Question: Does the paper fully disclose all the information needed to reproduce the main experimental results of the paper to the extent that it affects the main claims and/or conclusions of the paper (regardless of whether the code and data are provided or not)?
    \item[] Answer: \answerYes{} % Replace by \answerYes{}, \answerNo{}, or \answerNA{}.
    \item[] Justification: Yes, we provided a dedicated "implementation details" section in the appendix, which includes all the necessary information to reproduce our results. Moreover the code, dataset splits and pretrained checkpoints will be released upon publication.

\item {\bf Open access to data and code}
    \item[] Question: Does the paper provide open access to the data and code, with sufficient instructions to faithfully reproduce the main experimental results, as described in supplemental material?
    \item[] Answer: \answerYes{} % Replace by \answerYes{}, \answerNo{}, or \answerNA{}.
    \item[] Justification: Yes, we will release the code, dataset splits and pretrained checkpoints upon publication.

\item {\bf Experimental setting/details}
    \item[] Question: Does the paper specify all the training and test details (e.g., data splits, hyperparameters, how they were chosen, type of optimizer) necessary to understand the results?
    \item[] Answer: \answerYes{} % Replace by \answerYes{}, \answerNo{}, or \answerNA{}.
    \item[] Justification: Yes. In particular section \ref{sec:split} carefully explore the question of data splitting.

\item {\bf Experiment statistical significance}
    \item[] Question: Does the paper report error bars suitably and correctly defined or other appropriate information about the statistical significance of the experiments?
    \item[] Answer: \answerYes{} % Replace by \answerYes{}, \answerNo{}, or \answerNA{}.
    \item[] Yes, we applicable, we report the standard deviation over different random splitting of the datasets.

\item {\bf Experiments compute resources}
    \item[] Question: For each experiment, does the paper provide sufficient information on the computer resources (type of compute workers, memory, time of execution) needed to reproduce the experiments?
    \item[] Answer: \answerYes{} % Replace by \answerYes{}, \answerNo{}, or \answerNA{}.
    \item[] Justification: As explained in Appendix \ref{appendix:details} all experiment are conducted on a single V100 GPU and the execution time are also reported.
    
\item {\bf Code of ethics}
    \item[] Question: Does the research conducted in the paper conform, in every respect, with the NeurIPS Code of Ethics \url{https://neurips.cc/public/EthicsGuidelines}?
    \item[] Answer: \answerYes{} % Replace by \answerYes{}, \answerNo{}, or \answerNA{}.
    \item[] Justification: Yes, the research conducted in the paper conforms to the Code of Ethics.

\item {\bf Broader impacts}
    \item[] Question: Does the paper discuss both potential positive societal impacts and negative societal impacts of the work performed?
    \item[] Answer: \answerNA{} % Replace by \answerYes{}, \answerNo{}, or \answerNA{}.
    \item[] Justification: Metabolite Elucidation is a fundamental problem in metabolomics, which is a key technology for understanding biological systems and has applications in health, agriculture, and environmental science. The proposed method, MSAlign, has the potential to significantly improve the accuracy of metabolite identification from mass spectrometry data, which could accelerate research in these fields and lead to new discoveries. We do not identify any direct negative societal impacts of this work.
    
\item {\bf Safeguards}
    \item[] Question: Does the paper describe safeguards that have been put in place for responsible release of data or models that have a high risk for misuse (e.g., pre-trained language models, image generators, or scraped datasets)?
    \item[] Answer: \answerNA{} % Replace by \answerYes{}, \answerNo{}, or \answerNA{}.
    \item[] Justification: There is no risk for misuse of the data or models released in this paper, so we do not have any safeguards to put in place.

\item {\bf Licenses for existing assets}
    \item[] Question: Are the creators or original owners of assets (e.g., code, data, models), used in the paper, properly credited and are the license and terms of use explicitly mentioned and properly respected?
    \item[] Answer: \answerYes{} % Replace by \answerYes{}, \answerNo{}, or \answerNA{}.
    \item[] Justification: The authors have properly credited the creators of both models, benchmarks and datasets mentioned in the paper.

\item {\bf New assets}
    \item[] Question: Are new assets introduced in the paper well documented and is the documentation provided alongside the assets?
    \item[] Answer: \answerYes{} % Replace by \answerYes{}, \answerNo{}, or \answerNA{}.
    \item[] Justification: Upon publication, we will release the code, dataset splits and pretrained checkpoints. The code will be accompanied by documentation to facilitate its use by other researchers.

\item {\bf Crowdsourcing and research with human subjects}
    \item[] Question: For crowdsourcing experiments and research with human subjects, does the paper include the full text of instructions given to participants and screenshots, if applicable, as well as details about compensation (if any)? 
    \item[] Answer: \answerNA{} % Replace by \answerYes{}, \answerNo{}, or \answerNA{}.
    \item[] Justification: We did not conduct any crowdsourcing experiments nor research with human subjects.

\item {\bf Institutional review board (IRB) approvals or equivalent for research with human subjects}
    \item[] Question: Does the paper describe potential risks incurred by study participants, whether such risks were disclosed to the subjects, and whether Institutional Review Board (IRB) approvals (or an equivalent approval/review based on the requirements of your country or institution) were obtained?
    \item[] Answer: \answerNA{} % Replace by \answerYes{}, \answerNo{}, or \answerNA{}.
    \item[] Justification: We did not conduct any research with human subjects.

\item {\bf Declaration of LLM usage}
    \item[] Question: Does the paper describe the usage of LLMs if it is an important, original, or non-standard component of the core methods in this research? Note that if the LLM is used only for writing, editing, or formatting purposes and does \emph{not} impact the core methodology, scientific rigor, or originality of the research, declaration is not required.
    %this research? 
    \item[] Answer: \answerNA{} % Replace by \answerYes{}, \answerNo{}, or \answerNA{}.
    \item[] Justification: The core method development in this research does not involve LLMs beyond formating the paper and standard code auto-completion, which do not impact the core methodology, scientific rigor, or originality of the research.

\end{enumerate}

\end{document}